\documentclass[10pt,twocolumn,letterpaper]{article}

\usepackage[pagenumbers]{cvpr} %

\usepackage{graphicx}
\usepackage{amsmath}
\usepackage{amssymb}
\usepackage{booktabs}

\usepackage{xcolor}
\usepackage{graphicx}
\usepackage{float}
\usepackage{multicol}
\usepackage{multirow}
\usepackage{caption}
\usepackage{subcaption}
\usepackage{booktabs}
\usepackage{xparse}
\usepackage{xspace}
\usepackage{paralist}
\usepackage{wrapfig}
\usepackage{caption}
\usepackage{subcaption}
\usepackage{enumitem}
\usepackage{tabularx}
\usepackage{microtype}
\usepackage{comment}
\usepackage{cuted}
\usepackage{balance}

\usepackage{pifont}%
\newcommand{\cmark}{\ding{51}}%
\newcommand{\xmark}{\ding{55}}%

\usepackage[pagebackref,breaklinks,colorlinks]{hyperref}

\usepackage[capitalize]{cleveref}
\crefname{section}{Sec.}{Secs.}
\Crefname{section}{Section}{Sections}
\Crefname{table}{Table}{Tables}
\crefname{table}{Tab.}{Tabs.}

\definecolor{myOrange}{rgb}{1,0.5,0}
\definecolor{myBlue}{rgb}{0.1,0.2,0.8}
\definecolor{myGreen}{rgb}{0.1,0.8,0.1}
\definecolor{myRed}{rgb}{0.9,0.1,0.1}
\definecolor{myPurple}{rgb}{0.7,0.2,0.7}

\newcommand{\myparagraph}[1]{\vspace{0.75ex}\par\noindent \textbf{#1}}
\newcommand{\inst}[1]{\textsuperscript{#1}}

\newenvironment{packed_item}{
\begin{itemize}
  \setlength{\itemsep}{1pt}
  \setlength{\parskip}{2pt}
  \setlength{\parsep}{0pt}
}{\end{itemize}}

\def\cvprPaperID{8517} %
\def\confName{CVPR}
\def\confYear{2023}

\newcommand{\notation}[1]{\ensuremath{#1}\xspace}

\newcommand{\GANCriterion}{\notation{V}}
\newcommand{\Generator}{\notation{G}}
\newcommand{\Encoder}{\notation{E}}
\newcommand{\Discriminator}{\notation{D}}

\newcommand{\Dataset}{\notation{\mathcal{D}}}

\newcommand{\X}{\notation{x}}
\newcommand{\Y}{\notation{y}}
\newcommand{\Z}{\notation{z}}
\newcommand{\Ray}{\notation{\mathbf{r}}}

\newcommand{\LatentCode}{\notation{\mathbf{z}}}

\newcommand{\Noise}{\notation{n}} 

\newcommand{\Feature}{\notation{f}}

\newcommand{\FeatureLand}{\Feature_{\text{land}}}
\newcommand{\FeatureColor}{\Feature_{\text{color}}}
\newcommand{\FeatureColorSample}{\Feature_{\text{color}, i}}
\newcommand{\FeatureImage}{\Feature_{\text{im}}} %

\newcommand{\GeneratorBackground}{\Generator_{\text{sky}}}
\newcommand{\GeneratorUpsample}{\Generator_{\text{up}}}
\newcommand{\GeneratorLand}{\Generator_{\text{land}}}
\newcommand{\EncoderBackground}{\Encoder_{\text{clip}}}
\newcommand{\DiscriminatorFeature}{\Discriminator{\phi}}
\newcommand{\Reconstruction}{\Generator{\phi}}

\newcommand{\Image}{\notation{I}}
\newcommand{\Depth}{\notation{d}}
\newcommand{\Mask}{\notation{m}}
\newcommand{\Weight}{\notation{w}}
\newcommand{\Bilinear}{\notation{\beta}}

\newcommand{\ImageLR}{\Image_{\text{LR}}}
\newcommand{\ImageHR}{\Image_{\text{HR}}}
\newcommand{\DepthLR}{\Depth_{\text{LR}}}
\newcommand{\DepthHR}{\Depth_{\text{HR}}}
\newcommand{\MaskLR}{\Mask_{\text{LR}}}
\newcommand{\MaskHR}{\Mask_{\text{HR}}}
\newcommand{\ImageFull}{\Image_{\text{full}}}
\newcommand{\ImageSky}{\Image_{\text{sky}}}
\newcommand{\ImageSkyPano}{\Image_{\text{dome}}}

\newcommand{\MLP}{\notation{M}}
\newcommand{\Projection}{\notation{P}}
\newcommand{\Exp}{\notation{\mathrm{exp}}}

\begin{document}

\title{Persistent Nature: A Generative Model of Unbounded 3D Worlds}

\author{Lucy Chai\inst{1}, Richard Tucker\inst{2}, Zhengqi Li\inst{2}, Phillip Isola\inst{1}, Noah Snavely\inst{23}
\\[0.5em]
\inst{1}MIT \enspace \inst{2}Google Research \enspace \inst{3}Cornell Tech 
}

\maketitle

\begin{strip}
\vspace{-0.4in}
\centering
\includegraphics[width=1\textwidth]
{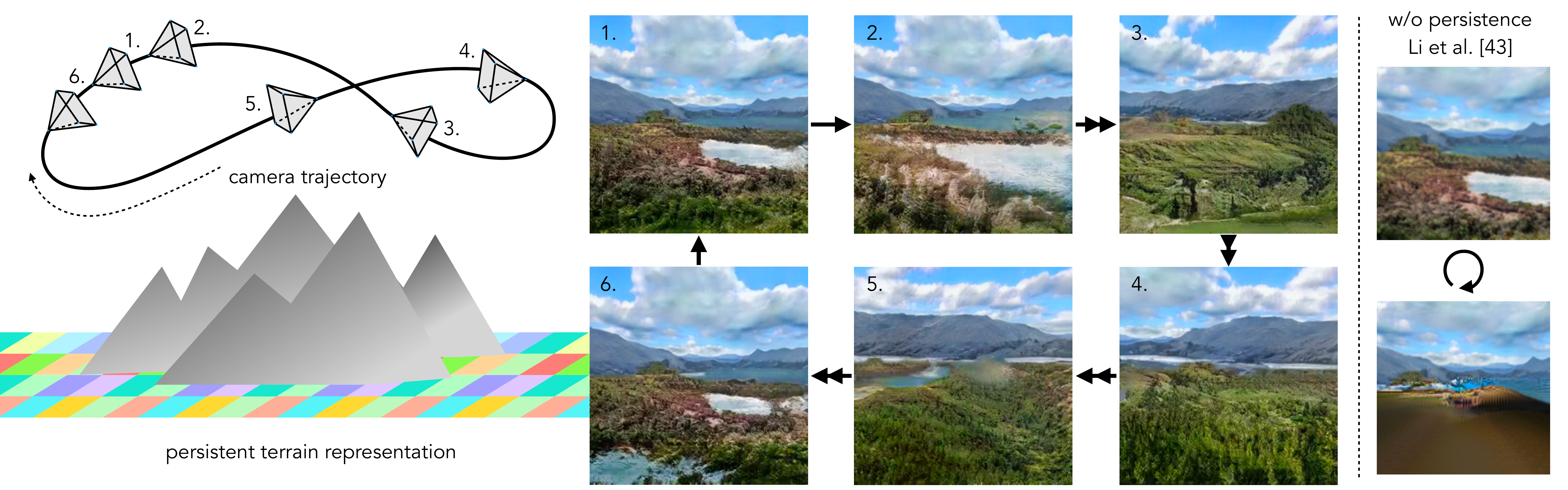}\vspace{-0.25cm}
\captionof{figure}{
\small Our approach enables unconditional synthesis of unbounded 3D nature scenes with a persistent scene representation (\textbf{left}), using a scene layout grid representing a large-scale terrain model (depicted above as the checkered ground plane). This representation enables us to generate arbitrary camera trajectories, such as the six numbered views shown along a cyclic camera path
(\textbf{center}).
The \emph{persistence} inherent to our representation stands in contrast to prior auto-regressive methods~\cite{li2022infinitenature} that do not preserve consistency under circular camera trajectories (\textbf{right}); 
while the two images shown on the right are at the start and end of a cyclic path, the terrain depicted is completely different. Our method %
is trained solely from unposed, single-view landscape photos.
}\label{fig:teaser}
\end{strip}

\begin{abstract}
\vspace{-5pt}
Despite increasingly realistic image quality, recent 3D image generative models often operate on 3D volumes of fixed extent 
with limited camera motions. 
We investigate the task of unconditionally synthesizing \emph{unbounded} nature scenes, enabling arbitrarily large camera motion while maintaining a persistent 3D world model. 
Our scene representation consists of an extendable, planar scene layout grid, which can be rendered from arbitrary camera poses via a 3D decoder and volume rendering, and a panoramic skydome.
Based on this representation, we learn a generative world model solely from single-view internet photos.
Our method enables simulating long flights through 3D landscapes, while maintaining global scene consistency---for instance, returning to the starting point yields the same view of the scene. 
Our approach enables scene extrapolation beyond the fixed bounds of current 3D generative models, while also supporting a persistent, camera-independent world representation that stands in contrast to auto-regressive 3D prediction models. Our project page: {\small \url{https://chail.github.io/persistent-nature/}}.
\end{abstract}

\section{Introduction}

Generative image and video models have achieved remarkable levels of realism, but are still far from presenting a convincing, explorable world.
Moving a virtual camera through these models---either in their latent space~\cite{harkonen2020ganspace,gansteerability,shen2020closed,brooks2022generating} or via explicit conditioning~\cite{drivegan}---is not like walking about in the real world.  
Movement 
is either very limited (for example, in object-centric models~\cite{chan2022eg3D}), or else camera motion is unlimited but quickly reveals the lack of a persistent world model.
Auto-regressive 3D synthesis methods exemplify this lack of persistence~\cite{liu2021infinite,li2022infinitenature}; parts of the scene may change unexpectedly as the camera moves,
and you may find that the scene is entirely different when returning to previous positions. The lack of spatial and temporal consistency can give the output of these models a strange, dream-like quality. 
In contrast, machines that can generate unbounded, persistent 3D worlds could be used to develop agents that plan within a world model~\cite{ha2018recurrent}, or to build virtual reality experiences that feel closer to the natural world, rather than appearing as ephemeral hallucinations~\cite{li2022infinitenature}.

We therefore aim to develop a unconditional generative model capable of generating unbounded 3D scenes with a persistent underlying world representation. We want synthesized content to move in a way that is consistent with camera motion, yet we should also be able to move arbitrarily far and still generate the same scene upon returning to a previous camera location, regardless of the camera trajectory. 

To achieve this goal, we model a 3D world as a \textit{terrain} plus a \textit{skydome}. The terrain is represented by a \textit{scene layout grid}---an extendable 2D array of feature vectors that acts as a map of the landscape. We `lift' these features into 3D and decode them with an MLP into a radiance field for volume rendering. The rendered terrain images are super-resolved and composited with renderings from the skydome model to synthesize final images. We train using a layout grid of limited size, but can extend the scene layout grid 
by any desired amount during inference, enabling unbounded camera trajectories.
Since our underlying representation is persistent over space and time, we can fly around 3D landscapes in a consistent manner. Our method does not require multiview data; each part of our system is trained from an unposed collection of single-view images using GAN objectives.

Our work builds upon two prior threads of research that 
tackle generating immersive worlds: 1) generative models of 3D data, and 2) generative models of infinite videos. Along the first direction are generators of meshes, volumes, radiance fields, etc (e.g., \cite{HoloGAN2019,chan2022eg3D,poole2022dreamfusion}). 
These models represent a consistent 3D world by construction, and excel at rendering isolated objects and bounded indoor scenes. Our work, in contrast, tackles the challenging problem of generating large-scale \emph{unbounded} nature scenes. %
Along the second direction are methods like InfiniteNature~\cite{liu2021infinite,li2022infinitenature}, which can indeed simulate visual worlds of infinite extent. 
These methods enable unbounded scene synthesis by predicting new viewpoints auto-regressively from a starting view. However, they do not ensure a persistent world representation; content may change when revisited. 

Our method aims to combine the best of both worlds,
generating boundless scenes (unlike prior 3D generators) while still representing a persistent 3D world (unlike prior video generative models). In summary:
\begin{packed_item}
    \item We present an unconditional 3D generative model for unbounded nature scenes with a persistent world representation, consisting of a terrain map and skydome.
    \item We augment our generative pipeline to support camera extrapolation beyond the training camera distribution by extending the terrain features. 
    \item Our model is learned entirely from single-view landscape photos with unknown camera poses. 
\end{packed_item}

\section{Related Work}

\noindent \textbf{Image and view extrapolation.}
Pioneering work by Kaneva~\etal~\cite{Kaneva_2010} proposed the task of infinite image extrapolation by using a large image database to perform classical 2D image retrieval, stitching, and rendering. 
More recently, various learning-based 2D image inpainting~\cite{hays2007scene,yu2018generative,yu2019free,liu2021pd,zhao2021large, suvorov2022resolution, li2022mat, saharia2022palette} and outpainting~\cite{wang2019wide,yang2019very,teterwak2019boundless,bowen2021oconet,lin2021infinitygan, cheng2022inout} methods have been developed.
These methods fill in missing image regions or expand the field of view by synthesizing realistic image content that is coherent with the partial input image. Beyond 2D, prior work has explored single-view 3D \emph{view extrapolation},
often by applying 2D image synthesis techniques within a 3D representation~\cite{wiles2020synsin, Shih3DP20, rockwell2021pixelsynth, hu2021worldsheet, rombach2021geometry, Li_2021_ICCV, jampani2021slide}.
However, these methods 
can only extrapolate content within a very limited range of viewpoints.

\smallskip
\noindent \textbf{Video generation.}
Video generation aims to synthesize realistic videos 
from different types of input. 
Unconditional video generation produces long videos often from noise input~\cite{tulyakov2018mocogan,munoz2021temporal,fox2021stylevideogan,skorokhodov2022stylegan, liu2021content, brooks2022generating, ge2022long}, 
while conditional video generation generates sequences by conditioning on one or a few images~\cite{vondrick2016generating,vondrick2017generating,wang2017predrnn,villegas2017decomposing,hsieh2018learning,lee2021revisiting, vondrick2016generating,finn2016unsupervised,denton2018stochastic,ye2019cvp,yu2022generating, koh2021pathdreamer}, or a text prompt~\cite{ho2022imagen, singer2022make}. 
However, applying these ideas in 3D requires supervision  from multi-view training data, and cannot achieve persistent 3D scene content at runtime, since there is no explicit 3D representation. 
Some recent work preserves global scene consistency via extra 3D geometry inputs such as point clouds~\cite{mallya2020world} or voxel grids~\cite{hao2021gancraft}. 
In contrast, our method synthesizes both the geometry and appearance of an entire world from scratch using a global feature representation to achieve consistent generated content.

\begin{figure*}[ht!] %
\centering
\includegraphics[width=1\textwidth]
{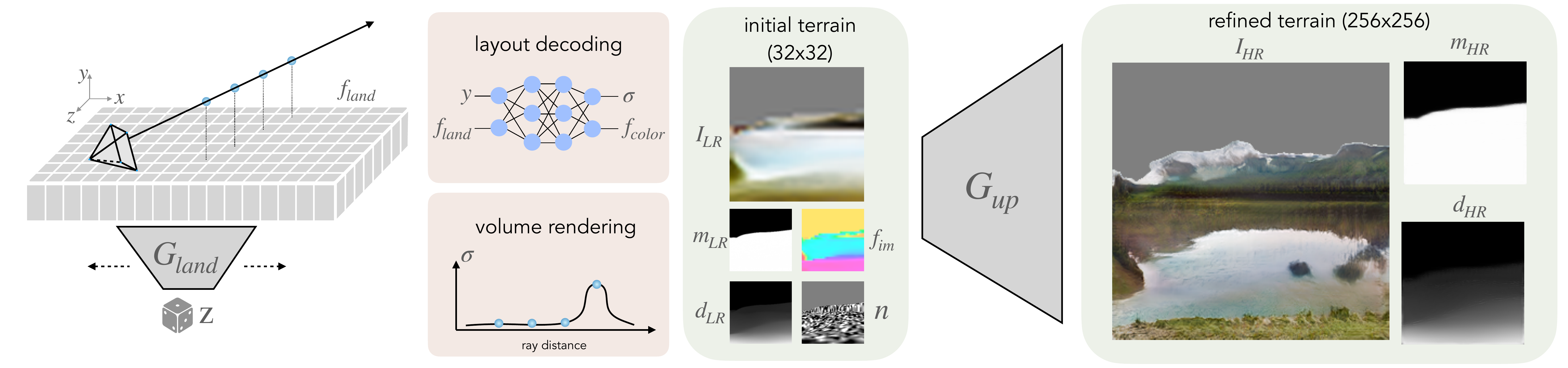}\vspace{-0.25cm}
\caption{\small \emph{Overview of scene layout decoding.} The layout generator $\GeneratorLand$ samples a random latent code to produce a 2D 
scene layout grid 
$\FeatureLand$ representing the shape and appearance of a terrain map, and which can be spatially extended using a grid of latent codes (see \S~\ref{sec:extension}).
To render an image from a given camera, 
sampled points along camera rays passing over the feature plane are decoded via an MLP into a color feature $\FeatureColor$ and density $\sigma$, which are then volume rendered. This produces a low-resolution image, mask, depth, image features, and a projected noise pattern, which are provided to a refinement network $\GeneratorUpsample$ to produce final image, mask, and depth outputs.
}\label{fig:schematic}\vspace{-3pt}
\end{figure*}

\smallskip
\noindent \textbf{Generative view synthesis.}
Novel view synthesis aims to produce new views of a scene from single~\cite{chen2019mono,tulsiani2018layer,niklaus20193d,single_view_mpi,shi2014light,wiles2020synsin,jang2021codenerf,Shih3DP20,Kopf-OneShot-2020,rombach2021geometry, yu2021pixelnerf} or 
multiple image observations~\cite{levoy1996light,zhou2018stereo,mildenhall2019local,flynn2019deepview,extremeview,lombardi2019neural,Riegler2020FVS,mildenhall2020nerf,wang2021ibrnet, muller2022instant, barron2022mip,yu_and_fridovichkeil2021plenoxels,shen2022sgam} by constructing a local or global 3D scene representation. However, most prior methods can only interpolate or extrapolate a limited 
distance from the input views, and do not possess a generative ability.

On the other hand, a number of generative view synthesis methods have been recently proposed utilizing neural volumetric representations~\cite{HoloGAN2019,schwarz2020graf,niemeyer2021giraffe,devries2021unconstrained,niemeyer2021campari,gu2021stylenerf,chan2022eg3D, rebain2022lolnerf,shorokhodov2022epigraf}. 
These methods can learn to generate 3D representations from 2D supervision, and have demonstrated impressive results on generating novel objects~\cite{poole2022dreamfusion}, faces~\cite{gu2021stylenerf, orel2022styleSDF, chan2022eg3D, Deng_2022_CVPR}, or indoor environments~\cite{ren2022look, devries2021unconstrained}. 
However, none of these methods 
can generate unbounded outdoor scenes due to lack of multi-view data for supervision, and due to the larger and more complex scene geometry and appearance that is difficult to model with prior representations. 
In contrast, our approach can generate globally consistent, large-scale nature scenes by training solely from unstructured 2D photo collections.

Our work is particularly inspired by recent perpetual view generation methods, including InfiniteNature~\cite{liu2021infinite} and InfiniteNature-Zero~\cite{li2022infinitenature}, which can generate unbounded fly-through videos of natural scenes, and are trained on nature videos or photo collections. 
However, these methods generate video sequences in an auto-regressive manner, and therefore
cannot achieve globally consistent 3D scene content. 
Our approach instead adopts a global scene representation that can be trained to generate consistent-by-construction and realistic novel views spanning large-scale scenes.
Concurrent works for scene synthesis InfiniCity~\cite{lin2023infinicity} and SceneDreamer\cite{chen2023scenedreamer} leverage birds-eye-view representations, while SceneScape~\cite{fridman2023scenescape} builds a mesh representation from text.

\section{Method}\label{sec:method}

Our scene representation for unbounded landscapes consists of two components, a \emph{scene layout grid} and a \emph{skydome}. 
The scene layout grid models the landscape terrain, and is a 2D grid of features defined on a ``ground plane.'' 
These 2D features are intended to describe both the height and appearance content of the terrain, representing the full 3D scene --- in fact, we decode these features to a 3D radiance field, which can then be rendered to an image (\S\ref{sec:scene-layout}). 
To enable camera motion beyond the training volume, 
we spatially extend the 2D feature grid to arbitrary sizes (\S\ref{sec:extension}). Because it is computationally expensive to generate and volume render highly detailed 3D content at the scale we aim for, we use an image-space refinement network that adds additional texture detail to rendered images (\S\ref{sec:refinement}).

The second scene component is a \emph{skydome} (\S\ref{sec:skydome}), which is a spherical (panoramic) image intended to model very remote content, such as the sun and sky, as well as distant mountains. The skydome is generated to harmonize with the terrain content described by the scene layout grid.

All the stages of our approach are trained with GAN losses (\S\ref{sec:training}). In what follows, we use the 3D coordinate convention that the ground plane is the $\textit{xz}$-plane, and the $y$-axis represents height above or below this plane. Generally, the camera used to view the scene will be positioned some height above the ground.

\subsection{Scene layout generation and rendering}\label{sec:scene-layout}

To represent a distribution over landscapes, we take a generative approach following the layout representation of GSN~\cite{devries2021unconstrained}. First, a 2D scene layout grid is 
synthesized from a sampled random noise code $\LatentCode$ passed to a StyleGAN2~\cite{karras2020analyzing} generator $\GeneratorLand$. This creates a 2D feature grid $\FeatureLand$, which we bilinearly interpolate to obtain a 2D function over spatial coordinates $\X$ and $\Z$:
\begin{equation}
\FeatureLand(\X, \Z) = \mathrm{Interpolate}(\GeneratorLand(\LatentCode), (\X, \Z))
\end{equation}
To define a full 3D scene, we need a way to compute the content at any 3D location $(\X, \Y, \Z)$. We define a multi-layer perceptron $\MLP$ that takes a scene grid feature, as well as the height $\Y$ of the point at which we want to evaluate the scene content. 
The outputs of $\MLP$ are the 2D-to-3D lifted feature $\FeatureColor$ and the density $\sigma$ at point $(\X, \Y, \Z)$:
\begin{equation}
\FeatureColor, \sigma = \MLP(\FeatureLand(\X, \Z), \Y).
\end{equation}
In this way, the 2D scene layout grid determines a radiance field over all 3D points within the bounds of the grid\cite{yu2021pixelnerf,devries2021unconstrained,sharma2022seeing}. That is, feature vectors in the grid encode not just appearance information, but also the height (or possibly multiple heights) of the terrain at their ground location.

To render an image from a desired camera pose, we cast rays $\Ray$ from the camera origin through 3D space, sample points $(\X, \Y, \Z)$ along them, and compute $\FeatureColor$ and $\sigma$ at each point. We then use volume rendering to composite $\FeatureColor$ along each ray into projected 2D image features $\FeatureImage$, a disparity image $\DepthLR$, and a sky segmentation mask $\MaskLR$. We form an initial RGB image of the terrain, $\ImageLR$, via a learned linear projection $\Projection$ of these image features. %
This process is depicted in the left half of Fig.~\ref{fig:schematic}, and is defined as:
\begin{equation}
\begin{split}
\FeatureImage(\Ray) &= \sum_{i=1}^N\Weight_{i}\FeatureColorSample, \quad
\DepthLR(\Ray) = \sum_{i=1}^N\Weight_{i}\Depth_i, \\
\MaskLR(\Ray) &= \sum_{i=1}^N\Weight_{i},
\hspace{14mm} \ImageLR = \Projection \FeatureImage, \\
\end{split}
\end{equation}
where $i\in\{1..N\}$ refers to the index of each sampled point along ray $\Ray$ in order of increasing distance from the camera, $\Depth_i$ is the inverse-depth (disparity) of point $i$, and weights $\Weight_{i}$ are determined
from the volume rendering equations used in NeRF~\cite{mildenhall2020nerf} (see supplemental).

We intend the mask $\MaskLR$ to distinguish sky regions (which will be empty and filled later using the skydome) from non-sky regions, and achieve this by training using segmented real images in which color and disparity for sky pixels are replaced with zero. Since to achieve zero disparity all weights along a ray must be zero (which also results in a zero-valued color feature), this approach encourages the generator to omit sky content.
However, 
while we find that the model indeed learns to generate transparent sky regions, land geometry can also become partially transparent. To counter this, we penalize visible decreases in opacity along viewing rays using finite differences of opacity $\alpha$:
\begin{equation}\label{eqn:geometry}
    \mathcal{L}_{\text{transparent}}(\Ray) = \sum_{i=2}^N \Weight_i \frac{\max(\alpha_{i-1}-\alpha_{i}, 0)}{\delta_i}.
\end{equation}

\subsection{Layout Extension}\label{sec:extension}

\begin{figure}[t!] %
\centering
\includegraphics[width=1\linewidth]
{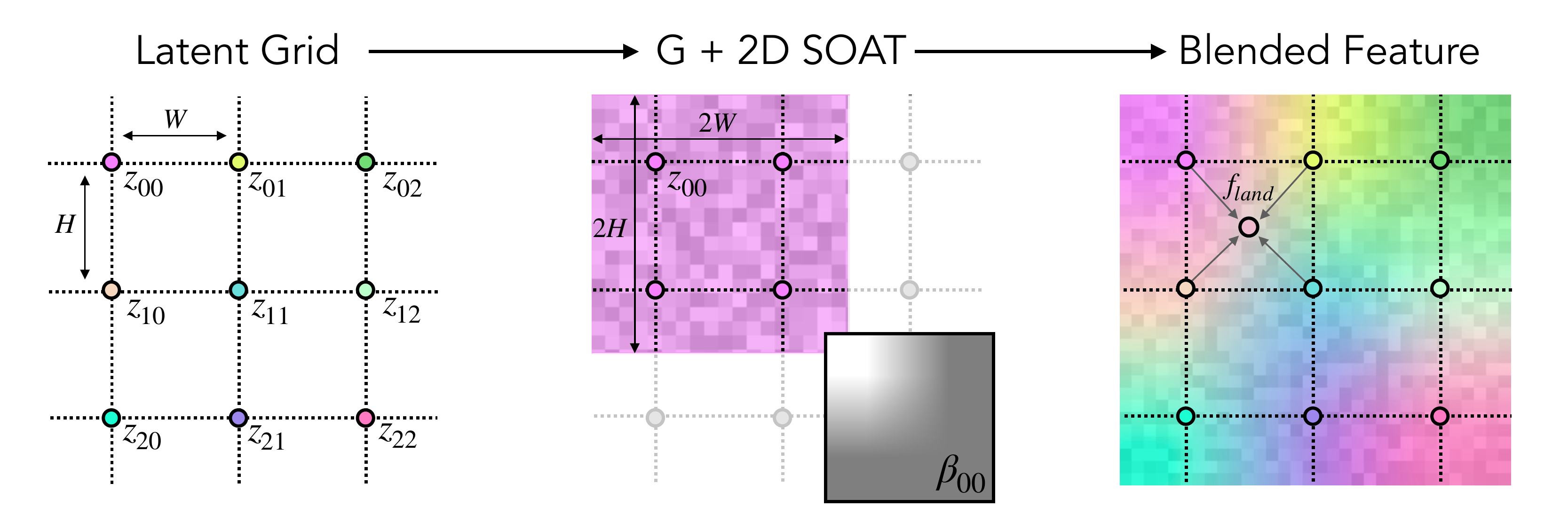}\vspace{-0.2cm}
\caption{\small \emph{Layout extension procedure.} To extend the layout at inference time, we sample noise codes \LatentCode in a grid arrangement. %
To smoothly transition between adjacent feature grids, we use the SOAT (StyleGAN of All Trades) procedure~\cite{chong2021stylegan} in 2D. Operating on a $2\times 2$ sub-grid, we apply each generator layer four times in fully convolutional manner over the entire sub-grid, each time conditioned on a different corner latent code \LatentCode, before multiplying by bilinear blending weights. This process is repeated for each layer of the generator and each sub-grid. Each $2\times 2$ sub-grid produces a $2H \times 2W$ feature grid, and sub-grids are blended together in an overlapping fashion to obtain an extended feature grid $\FeatureLand$ of arbitrary spatial size.
}\label{fig:stitching} \vspace{-3pt}
\end{figure}

While $\GeneratorLand$ creates a fixed-size feature grid, our objective is to generate geometry of arbitrary size, enabling long-distance camera motion at inference time. 
Hence, we devise a way to \textit{extend} the feature grid in the $\X$ and $\Z$ dimensions. We illustrate this process in Fig.~\ref{fig:stitching}, where we first sample noise codes $\LatentCode$ in a grid arrangement, where each $\LatentCode$ generates a 2D layout feature grid of size $H \times W$. To obtain a smooth transition between these independently sampled layout features, we generalize the image interpolation 
approach from SOAT (StyleGAN of all Trades)~\cite{chong2021stylegan} to two dimensions. We operate on $2\times2$ sub-grids and blend intermediate features from each layer of the generator as follows:
\begin{equation}\label{eqn:soat}
\begin{split}
\Feature_{k, l+1} & = \Generator_l(\Feature_l, \LatentCode_k);\quad k=\{00,~01,~10,~11\} \\
\Feature_{l+1} & = \sum_{k=\{00, 01, 10, 11\}}\Bilinear_k(\X, \Z)\Feature_{k, l+1}.
\end{split}
\end{equation}
For each of the four corner anchors $k$, we construct the modulated feature $\Feature_{k, l+1}$ by applying $\Generator_l$ (the $l$-th layer of $\GeneratorLand$) in a fully convolutional manner over the entire sub-grid. We then interpolate between the four feature grids using bilinear interpolation weights $\Bilinear_k(\X, \Z)$. By stitching these $2\times2$ sub-grids in an overlapping manner, we can obtain a scene layout feature grid of arbitrary size to use as $\FeatureLand$. Additional details are provided in the supplemental. 

\subsection{Image refinement}\label{sec:refinement}
Due to the computational cost of volume rendering, training the layout generator at higher resolutions becomes impractical. We therefore use a refinement network $\GeneratorUpsample$ to upsample the initial generated image $\ImageLR$ to a higher-resolution result $\ImageHR$, while adding  textural details (Fig.~\ref{fig:schematic}-right). We use a StyleGAN2 backbone for $\GeneratorUpsample$, replacing the earlier feature layers with feature output $\FeatureImage$ and the RGB residual layers with a concatenation of $\ImageLR$, $\DepthLR$, and $\MaskLR$.
To encourage the refined terrain image $\ImageHR$ to be consistent with the sky mask, the network also predicts a refined disparity map and sky mask 
for compositing with the skydome (see \S\ref{sec:skydome}): 
\begin{equation}\label{eqn:upsampler}
    \ImageHR, \DepthHR, \MaskHR = \GeneratorUpsample(\FeatureImage,\ImageLR, \DepthLR, \MaskLR).
\end{equation}
We compute a reconstruction loss between the initial and refined disparity and mask outputs, and  penalize $\GeneratorUpsample$ for producing gray sky pixels in $\ImageHR$ outside the predicted mask $\MaskHR$. Please see the  supplemental for more details. 

For fine texture details, StyleGAN2 also uses layer-wise spatial noise in intermediate generator layers (in addition to the global latent $\LatentCode$).
Using a fixed 2D noise pattern results in texture `sticking' as we move the camera~\cite{karras2021aliasfree}, but resampling it every frame reduces spatial coherence and removing it entirely results in convolutional gridding artifacts. To avoid these issues and improve spatial consistency, we replace the 2D image-space noise with projected 3D world-space noise, where the noise input to $\GeneratorUpsample$ is the projection of samples from a grid of noise, $\Noise$. This noise pattern is drawn from a standard Gaussian distribution defined on the ground plane at the same resolution of the layout features, which is then lifted into 3D and volume rendered along each ray $\Ray$:
\begin{equation} \label{eqn:noise}
\Noise(\Ray) = \sum_{i=1}^N\Weight_{i}\Noise(\X, \Z).
\vspace{-0.2cm}
\end{equation}

\begin{figure}[t!] %
\centering
\includegraphics[width=1\linewidth]
{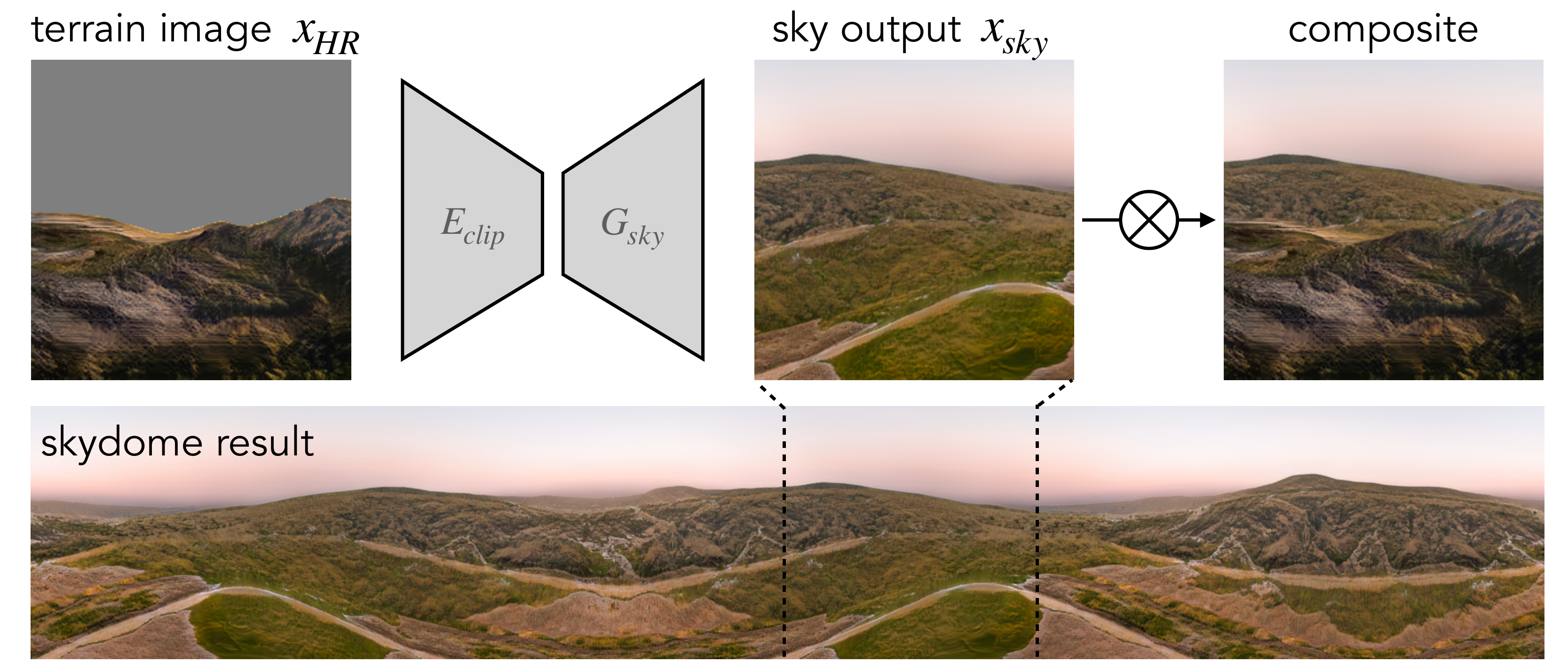}\vspace{-0.2cm}
\caption{\small \emph{Skydome generator.} Conditioned on the terrain image, the skydome generator $\GeneratorBackground$ synthesizes distant content (e.g., sky pixels and remote mountains) that is consistent with the generated terrain using encoder $\EncoderBackground$.  $\GeneratorBackground$ is conditioned on cylindrical coordinates which can be unwrapped to produce a panoramic skydome image.}\label{fig:skydome} \vspace{-3pt}
\end{figure}

\subsection{Skydome}\label{sec:skydome}
We model remote content (sky and distant mountains) separately with a skydome generator $\GeneratorBackground$ (Fig.\ref{fig:skydome}). This generator follows the StyleGAN3 architecture~\cite{karras2021aliasfree}, with a mapping network and synthesis network conditioned on cylindrical coordinates~\cite{chai2022anyresolution}. We adapt it by conditioning on the terrain output: we encode terrain images $\ImageHR$ using the pretrained CLIP image encoder $\EncoderBackground$~\cite{radford2021learning}, and concatenate this to the style-code output of the mapping network as input into $\GeneratorBackground$:
\begin{equation}
\begin{split}
    \ImageSky &= \GeneratorBackground(\mathrm{concat}(\EncoderBackground(\ImageHR), \mathrm{mapping}(\LatentCode))).
\end{split}
\end{equation}
Conditioning on the foreground terrain image encourages the skydome generator to generate a sky that is consistent with the terrain content. This model trains on single-view landscape images but can produce a full panorama at inference-time by passing in coordinates that correspond to a 360$^{\circ}$ cylinder.
The skydome is rendered to an individual camera viewpoint using camera ray directions, giving the skydome image $\ImageSkyPano$ which is then composited with the terrain image using the sky mask:
\begin{equation}
    \ImageFull = \ImageHR \odot \MaskHR + \ImageSkyPano \odot (1-\MaskHR).
\end{equation}

\begin{figure*}[ht!] %
\centering
\includegraphics[width=1\textwidth]
{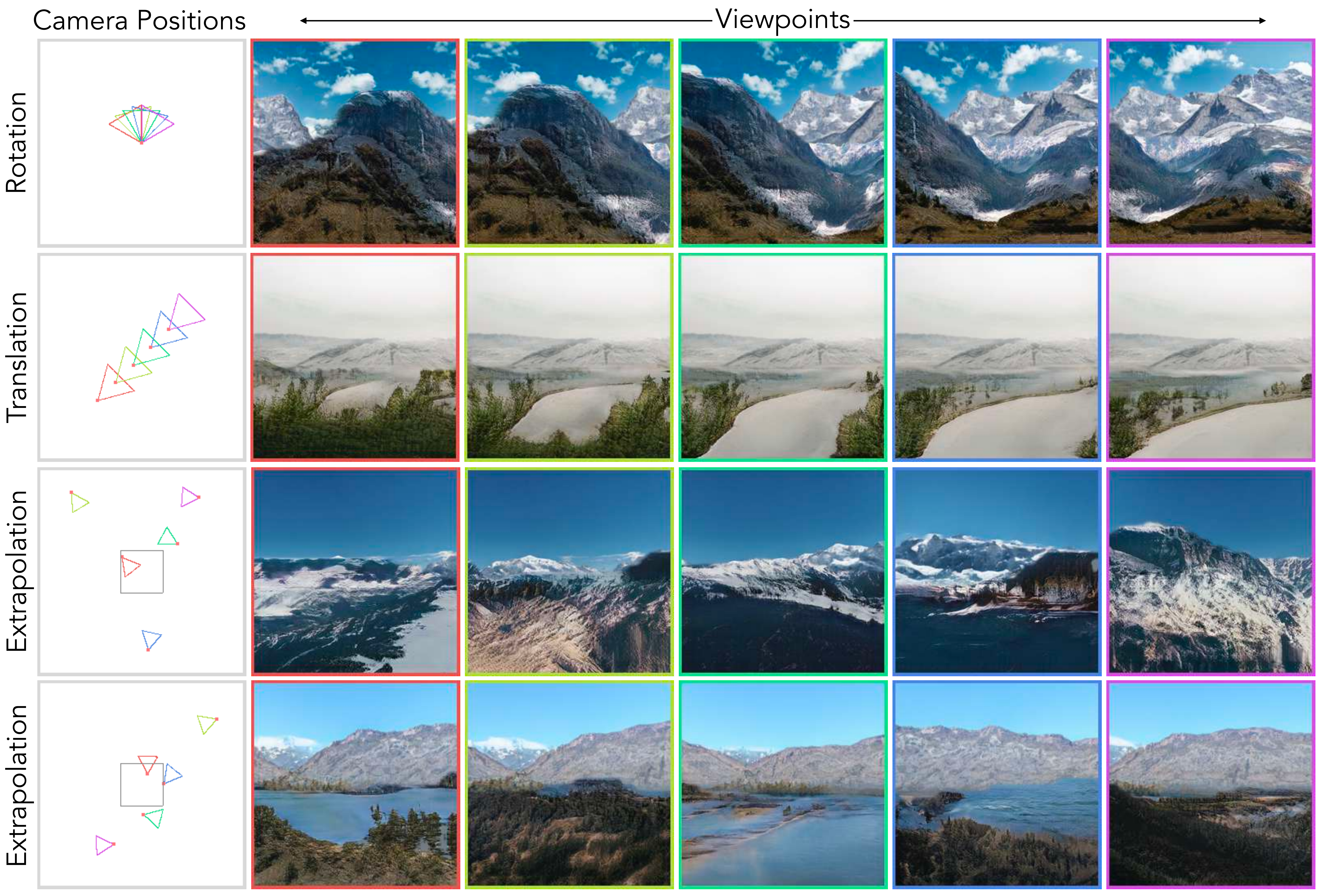}\vspace{-0.25cm}
\caption{\small \emph{Visualization of nearby and extrapolated camera motion.} Each row shows a set of sampled viewpoints, shown in an overhead view in the first column, and the corresponding rendered images in the other columns. Our model enables 3D-consistent view synthesis, visible under rotating or translating camera trajectories. We can also extrapolate the layout features at inference time, enabling camera motions outside of the training camera distribution (shown as a black square in the last two rows) with a consistent scene style.}\label{fig:qualitative}\vspace{-12pt}
\end{figure*}

\subsection{Training}\label{sec:training}
We train the layout generator (rendering at 32x32), refinement network (upsampling to 256x256), and skydome generator separately. To train the refinement network, we operate on outputs of the layout generator, freezing the weights of that model. For the skydome generator, we train using real landscape images, and apply it only to the outputs of the refinement network at inference time. We follow the StyleGAN2 objective~\cite{karras2020analyzing}, with additional losses for each training stage, architecture, and hyperparameters provided in the supplemental.

\myparagraph{Dataset and camera poses.} We train on LHQ~\cite{skorokhodov2021aligning}, a dataset of of 90K unposed, single-view images of natural landscapes. A number of LHQ images contain geometry that is not amenable to ``flying'', such as a landscape pictured through a window, or a closeup of trees. Therefore, we perform a filtering process on LHQ prior to training (see supplemental). We also obtain auxiliary outputs -- disparity and sky segmentation -- using the pretrained 
DPT~\cite{ranftl2021vision} model. 
Disparity and sky segmentation are used to construct the real image distribution in the GAN training phases.

After filtering, we use 56,982 images for training, and augment with horizontal flipping. During training we also need to sample camera poses. Prior 3D generators\cite{devries2021unconstrained,chan2021pi,chan2022eg3D,schwarz2020graf,orel2022styleSDF,gu2021stylenerf} either use ground-truth poses from a simulator, or assume an object-centric camera distribution in which the camera looks at a fixed origin from some radius. 
Because our dataset lacks ground truth poses, we first sample a bank of training poses uniformly across the layout feature grid with random small height offsets, and rotate such that the near half of the camera view frustum falls entirely within the layout grid. 
Since the aerial layout should not be specific to any given camera pose, we generate $\FeatureLand$ without any camera pose information, and then adopt the sampling scheme from GSN\cite{devries2021unconstrained} which samples a camera pose from the initial training pose bank proportional to the inverse terrain density at each camera position, to avoid placing the camera within occluding geometry.

\begin{figure*}[ht!] %
\centering
\includegraphics[width=1\textwidth]
{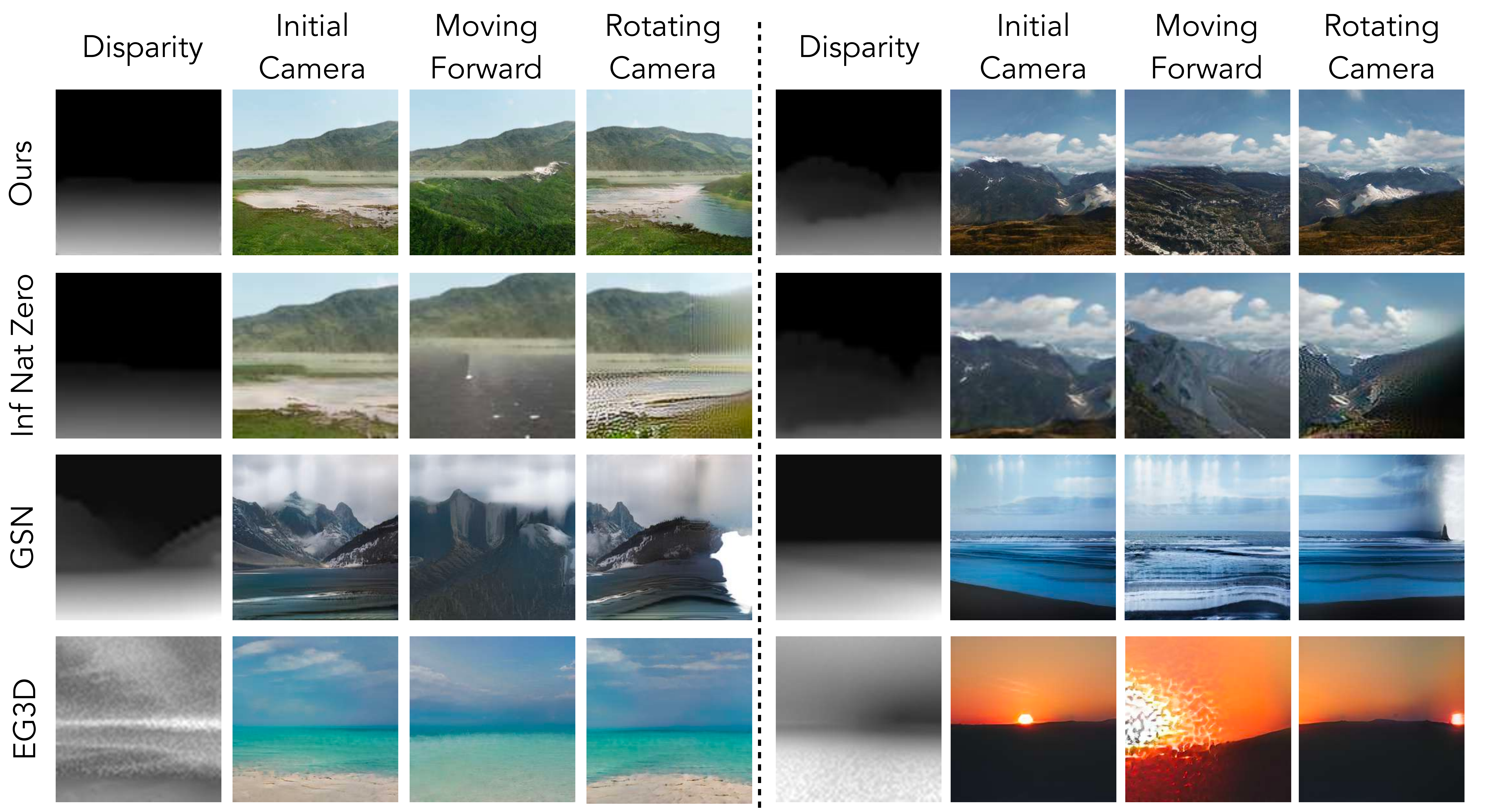}\vspace{-0.25cm}
\caption{\small \emph{Comparison to auto-regressive and bounded-volume 3D generative models.} Each row shows results for a given method on two generated scenes under different camera motion, along with a disparity map. Compared to InfiniteNature-Zero, our model enables long-range view synthesis by rendering a global scene description from different viewpoints, rather than auto-regressively predicting successive frames. 3D generative models like EG3D and GSN do not support view extrapolation on unbounded scenes. See our webpage for animated results. }\label{fig:models}\vspace{-10pt}
\end{figure*}

\begin{figure*}[ht!] %
\centering
\includegraphics[width=1\textwidth]{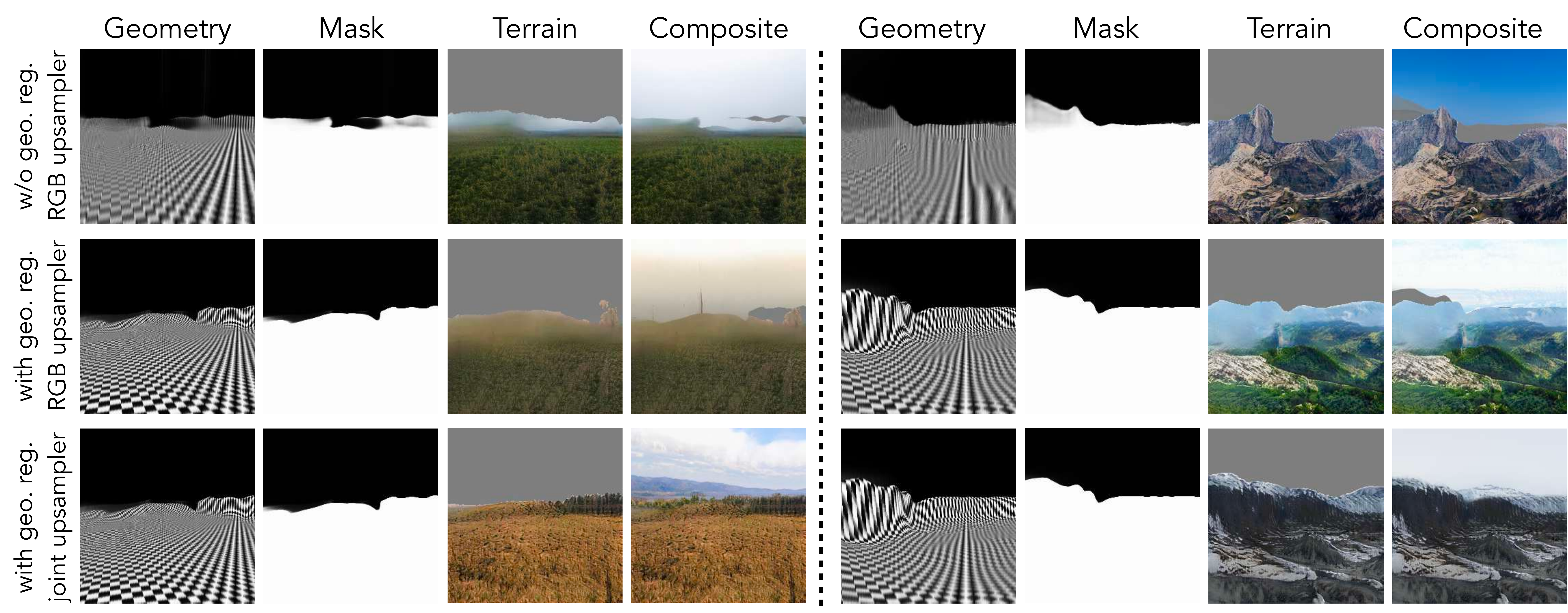}\vspace{-0.25cm}
\caption{\small \emph{Qualitative comparison of model variations.} Each row shows a model variant, visualizing generated geometry (as a rendered scene filled with a checkerboard pattern), sky mask, rendered terrain, and final image composite. 
(Top) Without geometry regularization, the model produces semi-transparent terrain. (Middle) Adding geometry regularization (Eqn.~\ref{eqn:geometry}) makes the terrain more solid, but there are inconsistencies between the terrain and mask prediction. (Bottom) Our full model uses geometry regularization and also adds a upsampler that operates on inverse-depth and sky mask inputs in addition to RGB (Eqn.~\ref{eqn:upsampler}) to discourage boundary effects between the terrain and sky. 
}
\label{fig:upsampler}\vspace{-10pt}
\end{figure*}

\section{Experiments}\label{sec:experiment}
Given its persistent scene representation and the extensibility of the its layout grid, our model enables arbitrary motion through a synthesized landscape, including long camera trajectories. 
We show sample outputs from our model under a variety of camera movements (\S~\ref{sec:expt_qualitative}); 
present qualitative and quantitative comparisons with alternate scene representations, including auto-regressive prediction models and unconditional generators defined for bounded or object-centric scenes (\S~\ref{sec:expt_scene_representation}); 
and investigate variations of our model to evaluate design decisions (\S~\ref{sec:expt_model_variations}).

\subsection{Persistent, unbounded scene synthesis} \label{sec:expt_qualitative}
Figure~\ref{fig:qualitative} shows example landscapes generated by our model with various camera motions. 
As the camera moves (by rotating and/or translating) the generated imagery changes in a way that is consistent with the underlying geometry, \eg hills move across the image or become closer. 
Extending the generated aerial feature grid allows us to place the camera \textit{outside} the distribution of training camera poses, while maintaining both geometric and stylistic consistency. 
As illustrated in Figure~\ref{fig:teaser} and our project page, the persistent and extendable layout features enables synthetic `flights' over large distances that can also return to a consistent starting point.

\subsection{Comparing scene representations}\label{sec:expt_scene_representation}
\begin{table}[t!]
  \centering
  \resizebox{1.0\linewidth}{!}{
  \begin{tabular}{lccccc}
    \toprule
     \multirow{2}{*}{\bf Model} & 
     \multirow{2}{*}{\bf Persistent} &
     \multirow{2}{*}{\bf Unbounded} & 
     \bf FID & \multicolumn{2}{c}{\bf Consistency}\\
     \cmidrule(lr){4-4}\cmidrule(lr){5-6}
     & &  & $C_\text{forward}$ & 1-step & cycle \\
    \midrule
    Inf Nat Zero~\cite{li2022infinitenature} & \xmark & \cmark & 28.15 & \bf 1.84 & 3.94 \\
    Ours (128px)  & \cmark & \cmark  & \bf 26.09 &  2.12 & \bf 0.00  \\
    \bottomrule
  \end{tabular}
  }
  \vspace{-0.25cm}
\caption{\small \emph{Comparison with InfiniteNature-Zero.} Using camera motions from InfiniteNature-Zero, we evaluate image quality as FID on 5K images after moving 100 steps forward ($C_\text{forward}$), one-step consistency as the L1 error when backwards warping one camera step, and cycle consistency as the L1 error between the original frame and the result after a pair of forward/backward steps. InfiniteNature-Zero is more consistent for a single step, but it has non-zero cyclic consistency error, and image quality degrades after repeated model applications. L1 values are multiplied by 100 throughout.}
\label{tab:quant_1}
\vspace{-0.5cm}
\end{table}

\begin{table}[t!]
  \centering
  \resizebox{1.0\linewidth}{!}{
  \begin{tabular}{lcccccc}
    \toprule
     \multirow{2}{*}{\bf Model} & 
     \multirow{2}{*}{\bf Persistent} &
     \multirow{2}{*}{\bf Unbounded} & 
     \multicolumn{3}{c}{\bf FID} & 
     \multirow{2}{*}{\bf Consistency} \\
     \cmidrule(lr){4-6}
     & & &  $C_\text{train}$ & $C_\text{forward}$ & $C_\text{random}$ &  \\
    \midrule
    GSN~\cite{devries2021unconstrained} & \cmark & \xmark & 29.95 & 50.22 & 45.48 & 12.80 \\ 
    EG3D~\cite{chan2022eg3D} & \cmark & \xmark & \bf 9.85 & 30.17 & 32.08 & \bf 3.01 \\
    Ours & \cmark & \cmark &  21.42 & \bf 26.67 & \bf 23.39 &  3.56 \\
    \bottomrule
  \end{tabular}
  }
  \vspace{-0.25cm}
\caption{\small \emph{Quantitative comparison to unconditional GANs.}
We evaluate image quality as FID on 5K images on (a) training camera poses $C_\textrm{train}$, (b) forward motion $C_\textrm{forward}$ (See Table~\ref{tab:quant_1}), (c) random camera poses $C_\textrm{random}$. One-step consistency error is measured
as the L1 error when backwards warping the result after one camera step to the initial frame, multiplied by 100. 
Once outside the training pose distribution our model generates better images than other methods, with consistency close to that of EG3D.}
\label{tab:quant_2}
\vspace{-0.5cm}
\end{table}

We compare our model with three state-of-the-art methods.
InfiniteNature-Zero is an auto-regressive method that, given an initial frame, generates successive frames sequentially by warping each image to the next based on depth~\cite{li2022infinitenature}. It allows for unbounded camera trajectories, but has no persistent world model. GSN~\cite{devries2021unconstrained} and EG3D~\cite{chan2022eg3D} are unconditional generative models: GSN uses a layout feature grid which is also the basis of our model, but focuses on bounded indoor scenes with ground-truth camera pose trajectories, while EG3D uses a tri-plane representation and primarily focuses on objects and portraits. These methods have persistent world models (feature grid and tri-plane representation) but do not allow for unbounded trajectories.

\myparagraph{Quantitative comparisons.} We evaluate image quality using FID~\cite{heusel2017gans}, and multi-view consistency using photometric error. To compare with InfiniteNature-Zero (Table~\ref{tab:quant_1}), we initialize with an image and depth map from our model, move the camera forwards using a forward motion trajectory from InfiniteNature-Zero, and evaluate image quality at a distance of 100 forward steps. 
Our model attains better FID, showing that it does not suffer from image degradation due to successive applications of an auto-regressive model. 
To compute one-step consistency error, we generate a new frame at a position equivalent to one forward step of InfiniteNature-Zero, warp it back to the original camera position using depth, and compute L1 error with the original frame in the overlapping region. Because InfiniteNature-Zero uses explicit warping as part of its model, it can achieve better one-step consistency, whereas our 2D upsampling operation is more susceptible to geometric inconsistency. 
We measure cyclic consistency error as the L1 error between the initial frame to the result after a step forward and back.
Because InfiniteNature-Zero lacks a persistent global representation, it has non-zero cyclic consistency error, whereas our model is fully consistent with zero cyclic consistency error.

To compare with the unconditional generative models GSN and EG3D, we compute FID on sets of output images corresponding to different distributions of camera positions: camera poses used in training which are intended to overlap with the layout, camera poses 100 steps forward from these mimicking InfiniteNature-Zero trajectories, and a uniform distribution of randomly oriented cameras 
over the layout grid. As seen in Table~\ref{tab:quant_2}, GSN is the least successful method when applied to this domain. EG3D generates high-quality images at training camera poses, but tends to represent the scene as floating nearby clouds with planar mountains at the edges of the volume (incorrect geometry).
Our method generalizes better to new camera positions. GSN has the highest one-step consistency error, while the consistency error of our model is close to that of EG3D (which relies less on 2D upsampling). In the supplemental, we experiment with an alternative architecture that builds on extendable triplane units with lower consistency error and faster rendering speed.

\myparagraph{Qualitative comparisons.} In Fig.~\ref{fig:models} we show example outputs of each model over forward-moving and rotating trajectories. Due to its auto-regressive nature, the quality of InfiniteNature-Zero's output degrades somewhat as the camera trajectory becomes longer. A more serious limitation is that, trained only on forward movement, it is unable to synthesize plausible views under camera rotation. GSN and EG3D also struggle with long camera trajectories, producing unrealistic outputs as the cameras approach the spatial limits of the training camera distribution. In the case of GSN applied to our setting, the results contain flickering and grid-like artifacts, which our projected noise (\S~\ref{sec:refinement}) mitigates.

\subsection{Model Variations}\label{sec:expt_model_variations}

To investigate individual components of our model, we separately evaluate variations of the layout generator and refinement network. 

\myparagraph{Layout generator.} The resolution of the scene layout grid and the number of samples per ray affect the quality of the volume-rendered output $\ImageLR$. As shown in Table~\ref{tab:layout}, higher resolution and more samples lead to the best image quality (FID computed on 32$\times$32 pixel images for speed, compared to segmented real images with gray sky pixels). To maximize the capacity of layout generation and rendering within computational limits we opt for a 256$\times$256 feature grid with 128 samples per ray.

\begin{table}[t]
  \centering
  \resizebox{0.7\linewidth}{!}{
  \begin{tabular}{lccc}
    \toprule
    \multirow{2}{*}{\bf Model} &
    \multirow{2}{*}{\shortstack[c]{\bf Samples\\\textbf{Per Ray}}} & 
    \multirow{2}{*}{\shortstack[c]{\bf Layout\\\textbf{Resolution}}} &
    \multirow{2}{*}{\shortstack[c]{\bf FID\\($\ImageLR$)}} \\
    & & & \\
    \midrule
    Low & 64 & 32 & 33.66 \\
    Medium & 128 & 32 & 32.02 \\
    High & 128 & 64 & 22.62 \\
    Full & 128 & 256 & \bf 16.06 \\ 
    \bottomrule
  \end{tabular}
}
\vspace{-0.25cm}
\caption{\small \emph{Variations on layout generation.} Higher feature grid resolution and more samples per ray yield the best results, bounded by computational limits. For speed, FID is computed on 5K samples rendered at 32$\times$32. \label{tab:layout}}
\vspace{-0.25cm}
\end{table}

\begin{table}[t]
  \centering
  \resizebox{1.0\linewidth}{!}{
  \begin{tabular}{lcccc}
    \toprule
    \multirow{2}{*}{\shortstack[c]{\bf Refinement\\\textbf{Output}}} &
    \multirow{2}{*}{\shortstack[c]{\bf Projected\\\textbf{Noise}}} &
    \multicolumn{2}{c}{\bf FID ($\ImageHR$)} &
    \multirow{2}{*}{\bf Consistency}\\
     \cmidrule(lr){3-4}
      & & $C_\text{train}$ & $C_\text{random}$ & \\
    \midrule
    $\ImageHR$ & \xmark & 26.30 & 27.08 & 5.08 \\
    $\ImageHR, \DepthHR, \MaskHR$ & \xmark & 23.75 & 27.25 & 5.81 \\
    $\ImageHR, \DepthHR, \MaskHR$ & \cmark & \bf 21.42 & \bf 23.39 & \bf 3.91 \\ 
    \bottomrule
  \end{tabular}
  }
  \vspace{-0.25cm}
  \caption{\small \emph{Variations on the refinement network.} We find refining not only the low-resolution image but also the depth and sky-mask improves image quality, but can lead to jittery results. The addition of projected noise into the upsampler results in smoother frames with lower consistency error. \label{tab:upsampler}}
  \vspace{-0.5cm}
\end{table}

\myparagraph{Refinement network.} Next, we investigate the refinement stage, which upsamples and refines the layout generator output. In our full model, the refinement network operates not only on RGB images but also on inverse-depth and sky mask (Eqn.~\ref{eqn:upsampler}), and uses projected noise for spatial consistency of texture detail (Eqn.~\ref{eqn:noise}). As shown in Table~\ref{tab:upsampler}, both help to improve our model's FID and consistency error. 

As shown in Fig.~\ref{fig:upsampler} (second row), upsampling only the RGB image $\ImageLR$ can lead to output that is inconsistent with the generated sky mask, leading to temporally unstable gaps in the final composited image. This figure also shows the effect of our geometric regularization (Eqn.~\ref{eqn:geometry}) in reducing unwanted transparency, especially in distant terrain.

\section{Discussion and conclusion}

\myparagraph{Limitations.}
A few drawbacks of our model include costly volume rendering limiting the resolution of $\ImageLR$, imperfect 3D consistency due to image-space refinement, and imperfect or repeating geometry decoded from the scene layout features. We elaborate in the supplemental. 

\myparagraph{Conclusion.}
We present an unconditional world generator for unbounded synthesis of persistent 3D nature scenes.
We build persistent world representation by modeling scene content with a spatially extendable layout feature grid which can be decoded via volume rendering to form a terrain image.
This rendered terrain is combined with a separate skydome, representing infinitely far content, to synthesize novel viewpoints supporting nearby and distant camera motions.
Altogether, our model enables 3D consistent image generation and view synthesis of unbounded scenes learned from single-view, unposed landscape photos.

\myparagraph{Acknowledgements.}
Thanks to Andrew Liu and Richard Bowen for the fruitful discussions and helpful comments. This project was part of an internship at Google.

\clearpage\newpage
{\small
\bibliographystyle{ieee_fullname}
\bibliography{egbib}
}

\clearpage
\appendix

In supplemental materials, we investigate an alternative 3D feature representation based on extendable triplane units~\ref{sec:sm_triplane}. We provide additional implementation details of our method in Section~\ref{sec:sm_method}, additional ablations in Section~\ref{sec:sm_experiment}, and 
we provide further discussion on our model in Section~\ref{sec:sm_discussion}.

\section{Extended Triplane Variation}\label{sec:sm_triplane}

\begin{figure}[ht!] %
\centering
\includegraphics[width=\linewidth]{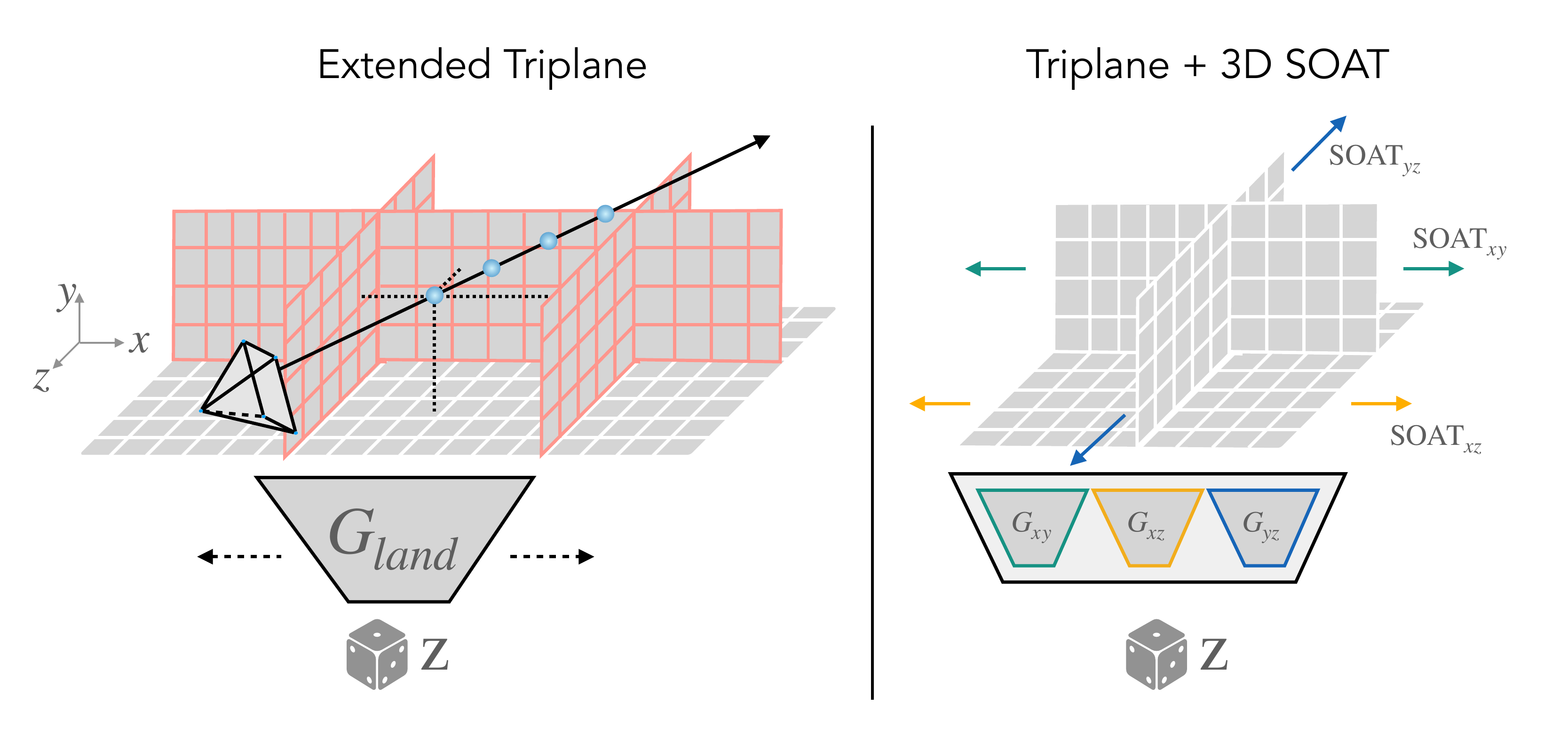}
\caption{\small \emph{Diagram of Extended Triplane Representation.} The extended triplane representation adds a sequence of orthogonal vertical feature planes outlined in pink in addition to the ground plane features outlined in white \textbf{(left)}. Each unit consists of a triplane representation~\cite{chan2022eg3D} generated from three independent generators -- $\Generator_{XY}$, $\Generator_{XZ}$, and $\Generator_{YZ}$ -- tied to the same latent code and mapping network \textbf{(right)}. At inference time, the features of each generator are stitched along the appropriate dimensions using the SOAT procedure~\cite{chong2021stylegan}.}
\label{fig:sm_expt_triplane}\vspace{-3pt}
\end{figure}

Instead of decoding the scene from a 2D layout feature grid and height of a 3D point above this layout plane, we also experiment with a model variation that adds vertical support planes parallel to the XY and YZ planes. Thus, the layout features are described by a 2D extended XZ layout feature grid, and sets of orthogonal support planes shown in pink in Fig.~\ref{fig:sm_expt_triplane}-left. Decoding a given 3D point projects the point to the XZ plane, the four nearest vertical planes (two parallel to XY and two parallel to YZ, which are weighted linearly according to the distance of the point from each plane). 

\begin{figure*}[ht!] %
\centering
\includegraphics[width=\linewidth]{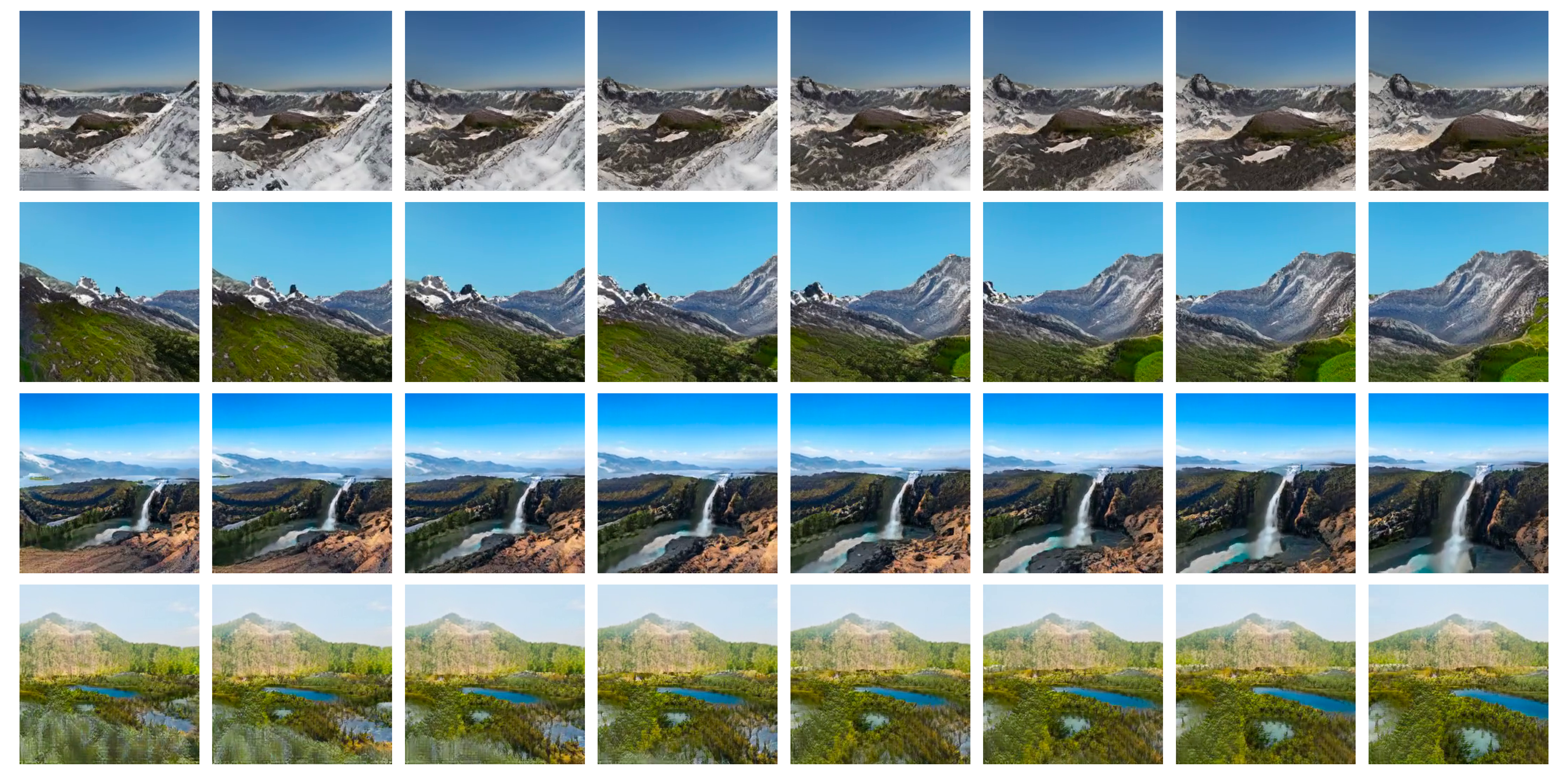}
\caption{\small \emph{Extendable Triplane Visualization.} Qualitative examples of rendering from the extendable triplane representation. This representation results in larger scene and geometry diversity compared to the layout feature representation, with improved 3D consistency. }
\label{fig:sm_expt_triplane_qualitative}\vspace{-3pt}
\end{figure*}

Qualitatively, the triplane model achieves more geometry diversity, with more mountainous terrain compared to the feature layout model. We attribute this to the additional support provided from the vertical feature planes. Additionally, the vertical feature planes allow for a lighter decoding network with higher neural rendering resolution, allowing for faster video rendering and improved temporal consistency (lower one-step consistency error) due to less reliance on a 2D upsampling operation. We show qualitative examples in Fig.~\ref{fig:sm_expt_triplane_qualitative} with video results on our project page, and quantitative evaluations in Tab.~\ref{tab:triplane_1}. Quantitatively, while this extended triplane variation does not output perform the layout model in terms of FID, we hypothesize that the FID may be impacted by two possible factors: first, this model requires inference-time camera height adjustment to avoid intersecting with increased complexity of the generated geometry, and second, interpolation between vertical feature planes qualitatively produces more muted colors compared to the real image distribution.

We also investigate the impact of using a projected 3D noise pattern as input into the extended triplane upsampler, with results in Tab.~\ref{tab:triplane_2}. While this improves FID and consistency in the layout representation, we find that the benefits of the projected noise are more limited in the extended triplane setting. Adding projected noise offers improvements in FID, but also a small increase in consistency error. Qualitatively, the model outputs are similar with and without the projected noise, perhaps attributed to decreased reliance on the upsampling operation.

\begin{table}[t!]
  \centering
  \resizebox{1.0\linewidth}{!}{
  \begin{tabular}{lccccc}
    \toprule
     \multirow{2}{*}{\bf Model} & 
     \multicolumn{3}{c}{\bf FID} & 
     \multirow{2}{*}{\bf Consistency} & 
     \multirow{2}{*}{\bf \shortstack[c]{\bf Render\\\textbf{Time (s)}}} \\
     \cmidrule(lr){2-4}
     &  $C_\mathrm{train}$ & $C_\mathrm{forward}$ & $C_\mathrm{random}$ &  & \\
    \midrule
    Extended Layout & \bf 21.42 & \bf 26.67 & \bf 23.39 & 3.56 &  8.49 \\ 
    Extended Triplane & 24.47 & 34.89 & 34.76 & \bf 2.29 & \bf 0.16 \\
    \bottomrule
  \end{tabular}
  }
\caption{\small \emph{Extended Layout vs. Extended Triplane}
While the extended layout representation presented in the main paper attains better image quality (lower FID scores), the extended triplane representation offers improved consistency (lower one-step consistency error) and dramatically faster video rendering (as the layout model requires supersampling for video smoothness). We hypothesize that inference-time camera adjustments and interpolation between vertical feature planes may negatively impact FID for the extended triplane model, despite its ability to generate more complex and diverse landscape geometry.
}
\label{tab:triplane_1}
\end{table}

\begin{table}[t!]
  \centering
  \resizebox{0.9\linewidth}{!}{
  \begin{tabular}{lcccc}
    \toprule
     \multirow{2}{*}{\bf Model} & 
     \multicolumn{3}{c}{\bf FID} & 
    \multirow{2}{*}{\bf Consistency} \\
     \cmidrule(lr){2-4}
     &  $C_\mathrm{train}$ & $C_\mathrm{forward}$ & $C_\mathrm{random}$ &  \\
    \midrule
    Without Noise & \bf 24.47 &	34.89 & 34.76 & \bf 2.29 \\ 
    With 3D noise & 25.31 &	\bf 33.30 & \bf 33.28 & 3.06 \\
    \bottomrule
  \end{tabular}
  }
\caption{\small \emph{Effect of 3D Projected Noise} Adding projected noise into the upsampler of the extendable triplane representation offers improvements in FID but is slightly more inconsistent, but still more consistent than the layout model.}
\label{tab:triplane_2}
\end{table}

\section{Additional Methodological Details}\label{sec:sm_method}

\subsection{Preprocessing}

\begin{figure*}[ht!] %
\centering

\begin{subfigure}[b]{0.48\textwidth}
      \includegraphics[width=\textwidth]{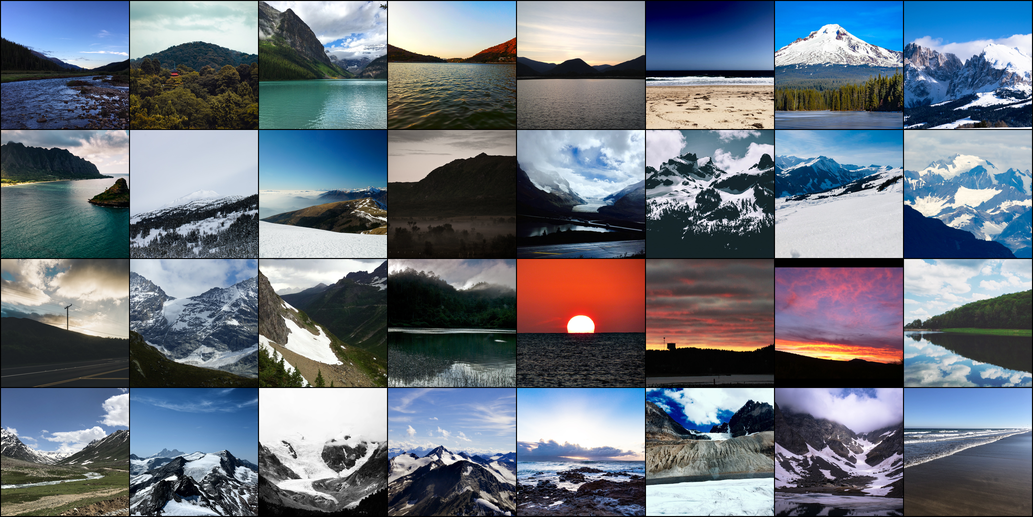}
      \caption{Training Images}
\end{subfigure}
\begin{subfigure}[b]{0.48\textwidth}
      \includegraphics[width=\textwidth]{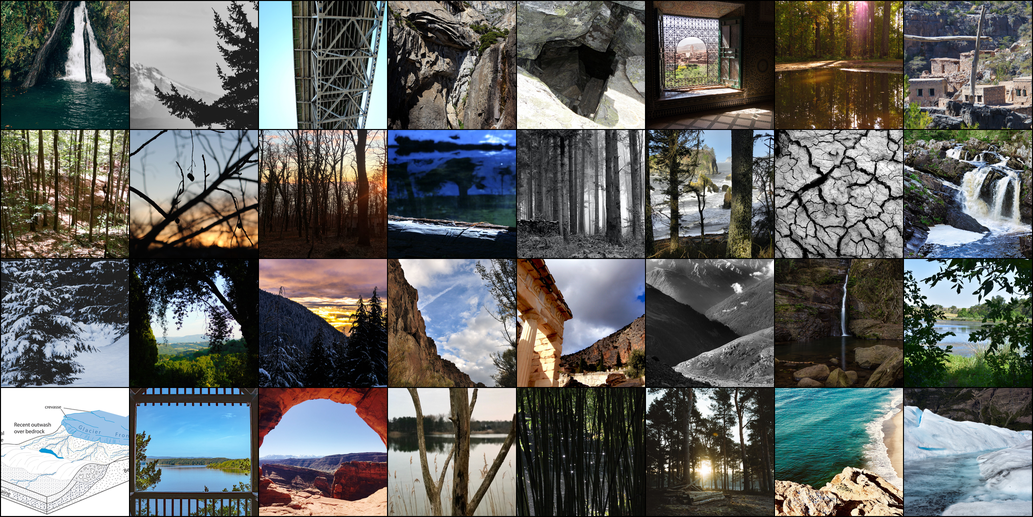}
      \caption{Removed Images}
\end{subfigure}
\caption{\small \emph{Result of dataset filtering.} The dataset filtering step (a) retains images that contain sufficient sky pixels near the top of the image, and (b) removes images that are not 
typical images of landscapes. These atypical images include images without sky pixels, or images with nearby occluding objects such as windows or trees. The filtering criteria is based on sky segmentation and disparity estimation obtained from DPT~\cite{ranftl2021vision}.}
\label{fig:sm_dataset}\vspace{-3pt}
\end{figure*}

\myparagraph{Dataset Filtering.} 
To remove images in the LHQ~\cite{skorokhodov2021aligning} dataset that contain occluding objects close to the camera, we apply filtering criteria to construct the training dataset. Using the segmentation output of DPT~\cite{ranftl2021vision}, we detect the sky region and boundaries of the resulting binary sky mask. As the segmentation results can include small regions with inconsistent labels (\eg small holes in the sky), we remove all bounded regions with area under 250 pixels to create a more unified sky mask. Next, using this segmentation mask we filter out images for which any of the following hold: (1) there are more than three bounded sky regions, (2) more than 90\% of the scene is not sky pixels, (3) more than 40\% of the upper one-fifth of the image is not sky pixels, and (4) less than 80\% of the lower quarter of the image is not sky pixels. The first three criteria are meant to filter out images that contain occluding structures (such as trees or windows) or images in which there is no sky region present. The fourth criteria is meant to filter out images taken from unusual camera angles (such as from underneath a bridge). Using the monocular depth prediction from DPT, we also remove images containing too many vertical edges: images are removed if the 99th percentile of the pixel-wise finite difference is greater than 0.05, which tends to be indicative of trees or man-made buildings. Fig.~\ref{fig:sm_dataset} shows examples of images that were retained for training, and those that were filtered out.

\myparagraph{Disparity Normalization.} Using the monocular depth prediction from DPT, we normalize the disparity values between 0 and 1 using the 1st and 99th percentile values per image. Next, we clip the minimum disparity for non-sky regions and rescale the disparity values to correspond to the near and far bounds used in volumetric rendering (see \S~\ref{sec:sm_layout}). We use 0.05 for our clip value and 1/16 for the scale factor; this means that after normalization, the disparity values for non-sky pixels range from 1/16 to 1. The disparity for the sky pixels is clamped at zero. 

\begin{figure*}[ht!] %
\centering
\begin{subfigure}[b]{0.32\textwidth}
      \includegraphics[width=\textwidth]{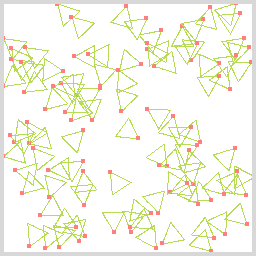}
      \caption{Training camera distribution}
\end{subfigure}
\begin{subfigure}[b]{0.32\textwidth}
      \includegraphics[width=\textwidth]{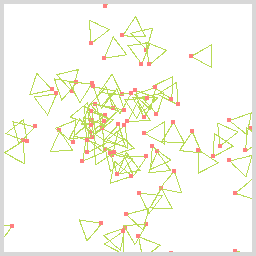}
      \caption{Cameras after forward motion}
\end{subfigure}
\begin{subfigure}[b]{0.32\textwidth}
      \includegraphics[width=\textwidth]{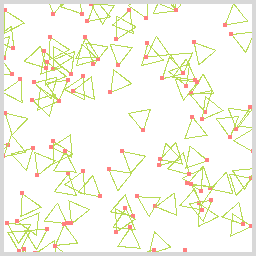}
      \caption{Random cameras}
\end{subfigure}
\caption{\small \emph{Illustration of camera distributions.} (a) Cameras used for training are sampled with a random translation uniformly over the scene layout feature grid, with rotation sampled to overlap with this feature grid. To evaluate view extrapolation, we (b) move the cameras forward a distance equivalent to 100 steps of InfiniteNature-Zero~\cite{li2022infinitenature}, corresponding to roughly halfway across the scene layout grid, or (c) randomly sample a random translation and random rotation.}
\label{fig:sm_cameras}\vspace{-3pt}
\end{figure*}

\myparagraph{Camera Poses.} We sample training camera poses with a random $(x, z)$ position within the layout grid, and a rotation such that the near half of the view frustrum lies entirely within the training grid. To simulate the forward motion of InfiniteNature-Zero~\cite{li2022infinitenature}, we move the camera forward a distance equivalent to 100 steps of InfiniteNature-Zero, corresponding to roughly half of the scene layout grid. To evaluate view extrapolation, we randomize the position and rotation of the cameras at inference time. These settings are illustrated in Fig.~\ref{fig:sm_cameras}. 

\subsection{Training and Implementation}

\subsubsection{Training objective}
Each stage of our model is trained following the StyleGAN2 objective~\cite{karras2020analyzing}, with a non-saturating GAN loss $\GANCriterion$ and $R_1$ regularization~\cite{mescheder2018training}:
\begin{equation}\label{eqn:gan}
\begin{split}
    &\GANCriterion(\Discriminator, \Generator(\LatentCode), \Image) = \Discriminator(\Image)-\Discriminator(\Generator(\LatentCode)), \\
    &R_1(\Discriminator, \Image) = || \nabla \Discriminator(x) || ^2, \\
    &\Generator = \arg \min_\Generator \max_\Discriminator
    \hspace{1mm} \mathbb{E}_{\LatentCode, \Image \sim \Dataset}
    \hspace{1mm} \GANCriterion(\Discriminator, \Generator(\LatentCode), \Image) + \\
    & \hspace{8mm} \frac{\lambda_{R_1}}{2} R_1(\Discriminator, \Image),
\end{split}
\end{equation}
where $\Generator,\Discriminator$ refer to the corresponding generator and discriminator networks at each training stage, and $x$ refers to real images sampled from dataset $\Dataset$. Additional auxiliary losses for each part of the model are described in the following sections.

\subsubsection{Layout Generator}\label{sec:sm_layout}

Our layout generator is based on the architecture from GSN~\cite{devries2021unconstrained}, which is comprised of two components: $\GeneratorLand$, which synthesizes the scene layout grid, and $\MLP$ which decodes the 2D layout feature into a 3D feature. 

The layout generator $\GeneratorLand$ follows StyleGAN2~\cite{karras2020analyzing}, which generates a $256 \times 256$ grid of features $\FeatureLand \in \mathbb{R}^{32}$. $\GeneratorLand$ contains three mapping layers and the maximum channel dimension is capped at 256; all other parameters are unchanged from StyleGAN2. 

The network $\MLP$ is modeled after the style-modulated MLP from CIPS~\cite{anokhin2021image}, containing eight layers with a hidden channel dimension of 256 and producing features $\FeatureColor \in \mathbb{R}^{128}$. The constant input to $\MLP$ is replaced with the $\Y$-coordinate (height above the ground plane), and the modulation input is the interpolated feature from $\FeatureLand$.

We adapt the rendering procedure of GSN to handle unbounded outdoor scenes. For volumetric rendering, we set the near bound to 1 and the far bound to 16, which corresponds to the scale factor used in disparity normalization during data preprocessing. Each scene layout feature has a unit width of $0.15$, such that the full width of the feature grid is $256 \times 0.15 = 38.4$, which is slightly over twice the far bound distance. We omit positional encoding from $\MLP$, as we found that including positional encoding yielded grid-aligned artifacts in generated images; we also omit the view direction input. Camera rays are sampled using $\mathrm{FOV}=60^{\circ}$ with linearly spaced sampling between the near bound and the far bound. We use inverse-depth (disparity) supervision rather than depth supervision so that we can represent content at infinite distances. This also encourages the terrain generator to create empty space in the sky content, which will be filled with the skydome generator.

We use the volumetric rendering equations from NeRF~\cite{mildenhall2020nerf}, in which the weights $\Weight_i$ of the $i$-th point along a ray depends on densities $\sigma$ which is predicted by multi-layer perceptron $\MLP$ and the distance between samples $\delta$:
\begin{equation}
\alpha_i = 1 - \Exp\left(-\sigma_i\delta_i\right),
\;\;
\Weight_{i} = \alpha_i \; \Exp\big(-\sum_{j=1}^{i-1}\sigma_j\delta_j\big).
\end{equation}

Our training procedure for the layout decoder follows that of GSN~\cite{devries2021unconstrained}, which provides the real RGB image $\Image_\textrm{RGB}$ and disparity $\Depth$ (obtained from DPT) to the discriminator $\Image=\{\Image_\textrm{RGB}, \Depth\}$, and also adds a reconstruction loss on real images using a decoder network $\Reconstruction$ on discriminator features $\DiscriminatorFeature$:
\begin{equation}
    \mathcal{L}_\textrm{rec} = (\Image - \Reconstruction (\DiscriminatorFeature (\Image)))^2.
\end{equation}
The full GAN objective follows Eqn.~\ref{eqn:gan} with weights $\lambda_{R_1} = 0.01$ and $\lambda_\textrm{rec} = 1000$, and we follow the optimizer settings from StyleGAN2 and train for 12M image samples. 

Because the layout decoder tends to generate semi-transparent geometry, which also causes unrealistic sky masks, we regularize the geometry following 
Eqn. \ref{eqn:geometry},
and add the sky mask into the discriminator. We finetune with this additional loss for 400k samples with $\lambda_{\mathrm{transparent}}$ which linearly increases from zero to $\lambda_{\mathrm{transparent}}=80$ over the finetuning procedure.

\subsubsection{Layout Extension}

\begin{figure}[ht!] %
\centering
\includegraphics[width=\linewidth]{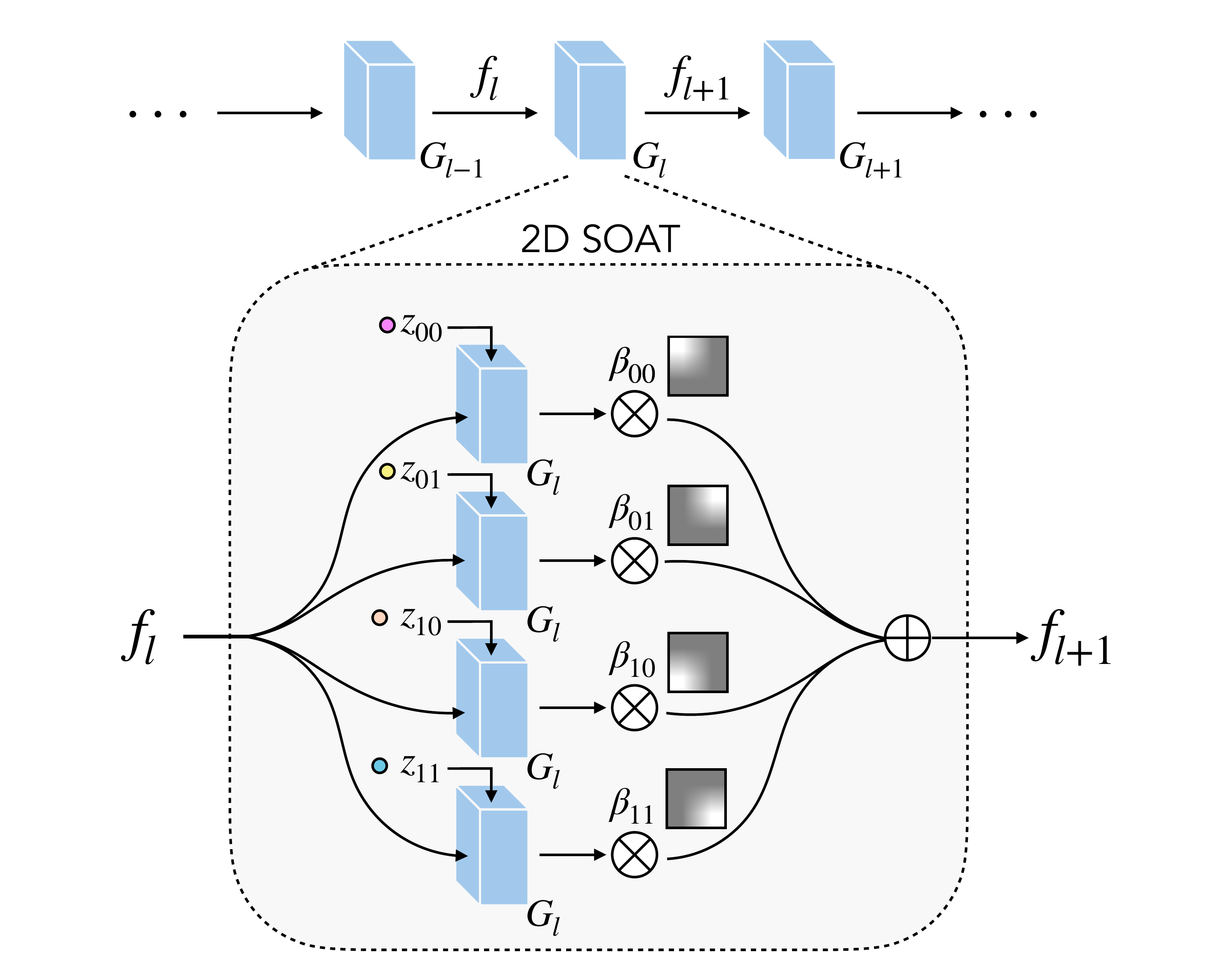}
\caption{\small \emph{Layout Extension.} We adapt the procedure in SOAT~\cite{chong2021stylegan} for 2D layout extension. Operating on each layer of the generator, we take the incoming feature grid $\Feature_l$, and construct the outgoing feature grid using the generator weights conditioned on each corner latent code $\LatentCode$ (the conditioning uses weight modulation on the mapping network outputs in StyleGAN2~\cite{karras2020analyzing}). Then, these four outgoing feature maps are multiplied with bilinear weights $\Bilinear$ and the result is summed, to obtain the blended feature for the next layer $\Feature_{l+1}$. }
\label{fig:sm_soat}\vspace{-3pt}
\end{figure}

We use the procedure of SOAT~\cite{chong2021stylegan} in two dimensions to smoothly transition between adjacent feature grids sampled from independent latent codes. SOAT proceeds by operating on 2x2 sub-grids and stitching each layer of intermediate features in the generator (Fig.~\ref{fig:sm_soat}). To start, we simply concatenate the StyleGAN constant tensors, to obtain a feature grid $\Feature_0$ of size $2H_0\times 2W_0$, where $H_0$ and $W_0$ are the height and width of the constant tensor. For each subsequent layer $\Feature_{l+1}$, we modulate the weights $G_l$ with each of four corner latent codes (after applying the mapping network to obtain the style-code) and apply it in a fully convolutional manner to $\Feature_l$, obtaining $\Feature_{k, l+1}$ of size $2H_l\times 2W_l$. Then, we multiply each of $\Feature_{k, l+1}$ with bilinear interpolation weights $\Bilinear$ and take the sum to obtain $\Feature_{l+1}$. This procedure is repeated for each layer of the generator, obtaining a an output feature grid of size $2H\times 2W$. To reduce the effect of padding, these output feature grids are tiled in an overlapping manner, with a 25\% overlap  on each side and with weights that linearly decay to zero away from the center of the tile.

\begin{figure*}[ht!] %
\centering
\begin{subfigure}[b]{0.45\textwidth}
      \includegraphics[width=\textwidth]{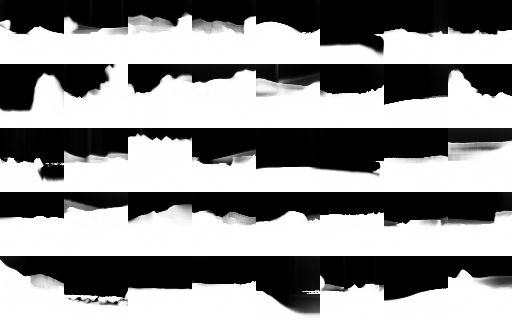}
      \caption{Accumulated ray density with separate skydome}
\end{subfigure}
\begin{subfigure}[b]{0.45\textwidth}
      \includegraphics[width=\textwidth]{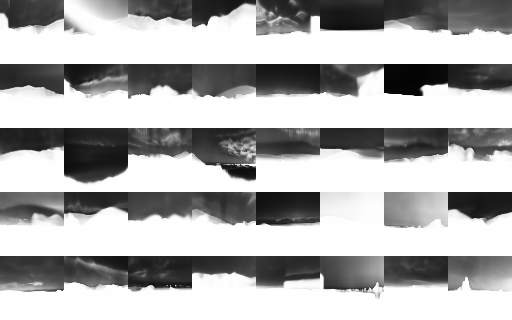}
      \caption{Accumulated ray density without separate skydome}
\end{subfigure}
\vspace{-5pt}
\caption{\small \emph{Training without a separate skydome.} We supervise the sky content with zero inverse-depth (infinite distance) to ensure that the camera does not intersect the sky as the layout features are extended. As such, we model content at infinite distances with a separate skydome model, such that the terrain model treats sky regions as empty space (left). Without the separated skydome, the model is forced to put sky content at finite distances leading to foggy, semi-transparent content near the camera (right). }
\label{fig:sm_expt_skydome}\vspace{-5pt}
\end{figure*}

\subsubsection{Refinement Network}

The refinement network $\GeneratorUpsample$ uses a truncated StyleGAN2 backbone, which replaces the feature input of the $32\times32$ block with the $32\times32$ rendered feature $\FeatureImage$ and initial image $\ImageLR$, depth $\DepthLR$, and sky mask $\MaskLR$. The skip connection of the upsampler takes in $\ImageLR$, $\DepthLR$, $\MaskLR$ and predicts $\ImageHR$, $\DepthHR$, and $\MaskHR$. Following the noise injection operation in StyleGAN2, we replace the image-space 2D noise tensor with our 3D-consistent projected noise 
(Eqn.~\ref{eqn:noise}).
This network uses two mapping layers, taking as input the style latent vector from $\GeneratorLand$. 

We add an additional objective to encourage consistency between the refined color pixels and the sky mask:
\begin{equation}\label{eqn:reg_upsampler}
\begin{split}
    \mathcal{L}_{\mathrm{consistency}} &= |\DepthHR - {\DepthLR}_\uparrow| + |\MaskHR - {\MaskLR}_\uparrow|, \\
    \mathcal{L}_{\mathrm{sky}} &= \exp(-20*\sum_{c}|\ImageHR[c]|) * \MaskHR. \\
\end{split}
\end{equation}
The loss $\mathcal{L}_{\mathrm{consistency}}$ encourages the high resolution depth and mask outputs to match their upsampled low resolution counterparts (this results in a smoother outcome compared to downsampling the high resolution outputs). The loss $\mathcal{L}_{\mathrm{sky}}$ encourages pixel colors to be nonzero (reserved for the gray sky color) when $\MaskHR=1$, by summing over the three channels $c$ of the predicted image $\ImageHR$; this is meant to encourage the RGB colors produced refinement network to be consistent with the mask and depth outputs.
The refinement network is trained with the GAN objective (Eqn.~\ref{eqn:gan}) with weights $\lambda_{R1}=4$, $\lambda_{\mathrm{consistency}}=5$, and $\lambda_\mathrm{sky}=100$, and the discriminator loss is applied only on the RGB images.  %

Due to the computational costs of volume rendering, we train the refinement network on $32\times32$ inputs to produce $256\times256$ outputs. For 30 fps video visualizations, we supersample the camera rays at 8x spatial density and apply depth-based filtering to the noise input to improve video smoothness; however all metrics in the paper are computed without supersampling for additional smoothness. 

We note that while StyleGAN3~\cite{karras2021aliasfree} is intended to resolve the texture sticking effect caused by the noise input in StyleGAN2, replacing $\GeneratorUpsample$ with a StyleGAN3 backbone resulted in worse image quality in our setting with FID 67.90, compared to FID 21.42 for our final model. 

\subsubsection{Skydome Generator}

The skydome generator takes as input the CLIP~\cite{radford2021learning} embedding of a single terrain image, and predicts a sky output that is consistent with the terrain. The generator architecture follows StyleGAN3~\cite{karras2021aliasfree} adapted with cylindrical coordinates to generate 360$^{\circ}$ panoramas~\cite{chai2022anyresolution}. 

For the terrain input, we take the filtered LHQ dataset and select the non-sky pixels with normalized disparity greater than 1/16 (this leaves some background mountains to be predicted). We follow the training procedure from~\cite{chai2022anyresolution} with a few adaptations. In addition to concatenating the CLIP embedding of the terrain image to the style-code, the generated sky is composited with the terrain input prior to the discriminator with 50\% probability, which is compared to full RGB images from LHQ. The 50\% compositing behavior ensures that the bottom of the generated skydome can still appear realistic (when unmasked), while also matching provided terrain image (when masked). This portion is trained with the $\lambda_{R1}=2$ in the GAN objective (Eqn.~\ref{eqn:gan}), with randomly sampled cylindrical coordinates and a cross-frame discriminator applied to the boundary of two adjacent frames.

\subsection{Extendable Triplane Implementation}

To construct the extendable triplane representation, we modify the triplane model from EG3D~\cite{chan2022eg3D} to generate three planes from independent synthesis networks $\Generator_{XY}$, $\Generator_{XZ}$, and $\Generator_{YZ}$, tied to the same latent code and mapping network. Similar to our layout feature model, we train the terrain generator on sky-segmented images and disparity maps as input into the low-resolution discriminator to help the model learn geometry. The upsampler portion of this model and the training procedure is the same as EG3D, using $\lambda_{R1}=10$. To prevent the model from rendering the segmented sky color (we use white for the sky color, following the background color of NeRF~\cite{mildenhall2020nerf}), we finetune the model penalizing for white pixels when the sky mask is one:
\begin{equation}
\mathcal{L}_{\mathrm{sky}} = \exp(-5*\sum_{c}(\ImageLR[c]-1) * \MaskLR. \\
\end{equation}
The finetuning operation is performed for 400K samples with $\lambda_{\mathrm{sky}}$ increasing from zero to 40 during training.
At inference time, we perform SOAT~\cite{chong2021stylegan} feature stitching to each generator along the appropriate dimensions to obtain the extended triplane representation. As the skydome model does not train on generated images, we use the same skydome model as before. We use 50 randomly sampled camera poses for training, which improves the geometry diversity (more mountainous terrain) the compared to using 1K random training poses.

\section{Additional Experiments}\label{sec:sm_experiment}

\subsection{Training without a separate skydome}

Modelling faraway content separately is a common strategy in unbounded scene-reconstruction~\cite{barron2022mip,hao2021gancraft}.
To ensure that we cannot intersect the skydome as we arbitrarily extend the layout features, we use zero inverse-depth for sky pixels, which can only render a solid color as the weight of all points along the ray must be zero to obtain zero inverse-depth. In this experiment, we train $\ImageLR$ using the same training strategy as our final $\ImageLR$ model, but instead supervise with full RGB images rather than sky-segmented RGB images. This corresponds to training the model without a separate skydome. We find that without the separate skydome, the model learns incorrect geometry, as it is forced to place some density at finite distances in order to render content in the sky to match the training distribution. Figure~\ref{fig:sm_expt_skydome} shows the difference in ray accumulations from models trained with the skydome (prior to opacity regularization) and without the skydome. The model without the skydome places semi-transparent content in the sky region, which creates a fog-like effect when moving the camera throughout the landscape. 

\subsection{Changing the number of sampled cameras}

We train our model using a set of one thousand cameras with randomly sampled translations within the layout feature grid, and rotations such that the camera view frustum overlaps with the feature grid. However, one limitation of this training strategy is that we find the model can learn repeating geometry, such that the rendered disparity map may look similar when sampling different random latent codes at the same camera position, despite the pixel color values being different. We hypothesize that the diversity of camera poses sampled during training may obscure the repeating geometry effect from the discriminator, as images sampled from different camera poses will appear different in terms of both color and geometry. %

\begin{figure*}[ht!] %
\centering
\begin{subfigure}[b]{0.45\textwidth}
      \includegraphics[width=\textwidth]{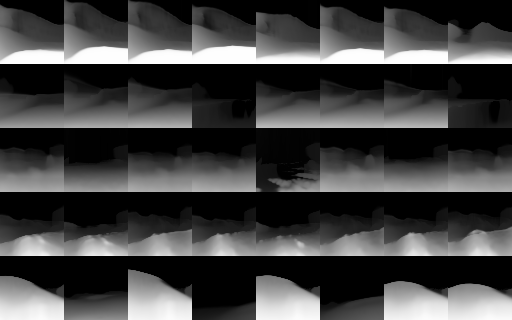}
      \caption{1K training cameras; training poses}
\end{subfigure}
\begin{subfigure}[b]{0.45\textwidth}
      \includegraphics[width=\textwidth]{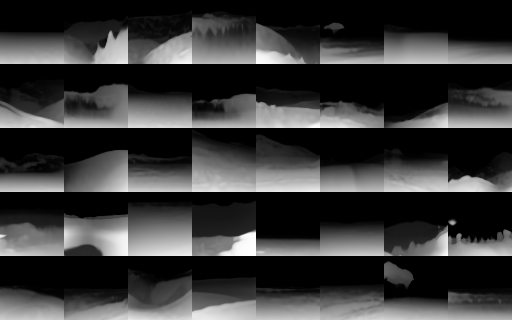}
      \caption{5 training cameras; training poses}
\end{subfigure}
\begin{subfigure}[b]{0.45\textwidth}
      \includegraphics[width=\textwidth]{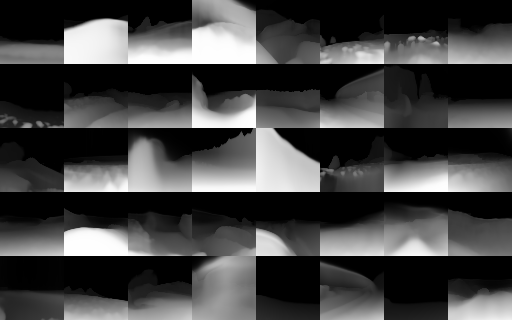}
      \caption{1K training cameras; independent test poses}
\end{subfigure}
\begin{subfigure}[b]{0.45\textwidth}
      \includegraphics[width=\textwidth]{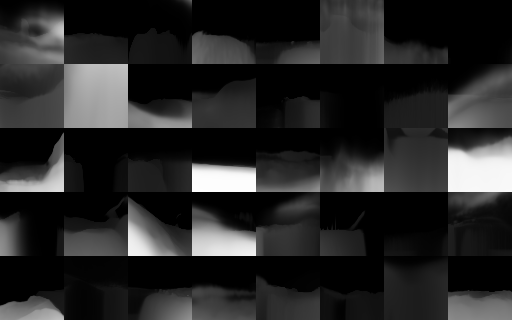}
      \caption{5 training cameras; independent test poses}
\end{subfigure}
\vspace{-5pt}
\caption{\small \emph{Adjusting the set of training cameras.} We plot disparity maps corresponding to training with one thousand cameras, and five cameras. (a) With our default setting of one thousand training cameras with camera origins uniformly sampled over the layout feature grid, we find that the model can learn repeating geometry, such that the disparity map generated from the same pose but different latent codes tends to look similar (each row corresponds to the same pose), despite the RGB colors appearing different. (b) With fewer training cameras, the models learns more diversity in the rendered geometry, where again each row corresponds to the same camera pose. (c \& d) However, the model trained with one thousand cameras generalizes better to an independent set of cameras, whereas the model trained with five cameras has a greater frequency to put holes in the decoded landscape (evidenced by completely black disparity maps, or disparity maps that have no nearby content and thus are darker overall) or regions of solid content without sky (evidenced by disparity maps that do not fade to black near the top of each image). We use one thousand training cameras as our default setting, but a more optimal setting may involve fewer training cameras, while still ensuring adequate coverage over the feature grid.
}
\label{fig:sm_expt_cameras}\vspace{-5pt}
\end{figure*}

To investigate this effect, we train another model using only five camera poses during training. The disparity maps per camera pose show more diversity in this setting, however we find that this setting results in ``holes'' and incorrect geometry in the landscape when moving the camera away from the training poses, illustrated in Figure~\ref{fig:sm_expt_cameras}. We use one thousand training cameras as our default setting, but a more optimal setting may involve fewer training cameras, while still ensuring adequate coverage over the feature grid.

\balance

\section{Discussion}\label{sec:sm_discussion}

A limiting factor of our method is the reliance on a volume rendering operation to decode the 2D layout feature grid into a 3D feature at each sampled point along the ray. Due to this operation, the rendered output $\ImageLR$ can only be trained at low resolution (32x32), and does not learn to generate detailed textures. (In contrast to NeRF-style models which can use per-ray supervision, we must render a complete image as an input for the discriminator.) We rely on a refinement module to upsample the result and add additional textures, but any refinement in image space is prone to losing 3D consistency.  Our extended triplane variation reduces the computational expense of volume rendering by reducing the capacity of the decoder MLP and increasing the capacity of the feature representation, thus allowing for neural rendering at 64x64 resolution (we find that geometry degrades at higher resolutions) and decreasing reliance on the upsampler. While we did not find improvements when training on rendered patches, improved patch sampling techniques could help in adding more detail to the rendered result~\cite{shorokhodov2022epigraf}.

As our model does not have explicit 3D or aerial supervision, we find that it may generate unnatural or repeating geometry. This can appear as thin mountains, sloping water, or hills of a similar shape but different appearance when sampling from different random noise codes.  

\end{document}